\documentclass[lettersize,journal]{IEEEtran}
\usepackage{amsmath,amsfonts}
\usepackage{fix-cm}
\usepackage{algorithmic}
\usepackage{algorithm}
\usepackage{array}
\usepackage[caption=false,font=normalsize,labelfont=sf,textfont=sf]{subfig}
\usepackage{textcomp}
\usepackage{tabularx}
\usepackage{stfloats}
\usepackage{url}
\usepackage{verbatim}
\usepackage{graphicx}
\usepackage{cite}
\usepackage{multirow}
\usepackage{silence}
\usepackage{booktabs}
\usepackage{enumitem}
\usepackage[normalem]{ulem}
\usepackage{utfsym}
\usepackage{marvosym}
\usepackage{
  color,
  float,
  epsfig,
  wrapfig,
  graphics
}
\definecolor{gray}{gray}{0.5}
\definecolor{revisionorange}{RGB}{230,120,0}
\definecolor{revisionblue}{RGB}{0,84,166}
\newcommand{\projectname}{\textsc{BadReward}}
\newcommand{\greencheck}{\textcolor{green}{\checkmark}}
\newcommand{\redcross}{\textcolor{red}{$\times$}}
\newcommand{\graydash}{\textcolor{gray}{\rule{0.5em}{0.4pt}}} 
\newcommand{\red}[1]{{\color{red} #1}}

\newcommand{\partitle}[1]{\smallskip \noindent \textbf{#1.}}

\begin{document}

\title{The Alignment Illusion: How \projectname{} Corrupts \\ Text-to-Image Generation}

\author{Kaiwen Duan, Hongwei Yao, Yufei Chen, Ziyun Li, Tong Qiao, Zhan Qin, and Cong Wang~\IEEEmembership{Fellow,~IEEE}

\thanks{This research is supported in part by the ``Pioneer'' and ``Leading Goose'' R\&D Program of Zhejiang (Grant No. 2026C01020), the Guangdong and Hong Kong Universities ``1+1+1'' Joint Research Collaboration Scheme, the National Natural Science Foundation of China (Grant No. 62441238, 62472135, U2441240 ``Ye Qisun'' Science Foundation), the Zhejiang Province Key R\&D Project (No. 2025C04002), and the Kunpeng--Ascend Science and Education Innovation Excellence/Incubation Center.}
\thanks{Kaiwen Duan and Zhan Qin are with Zhejiang University, Hangzhou, China.
Hongwei Yao, Yufei Chen, and Cong Wang are with City University of Hong Kong, Hong Kong, China.
Ziyun Li is with KTH Royal Institute of Technology, Stockholm, Sweden.
Tong Qiao is with Hangzhou Dianzi University, Hangzhou, China.
Zhan Qin and Hongwei Yao are also with Hangzhou High-Tech Zone (Binjiang) Institute of Blockchain and Data Security, Hangzhou, China.}
\thanks{This paper has supplementary downloadable material available at http://ieeexplore.ieee.org, provided by the author. The material includes reproducibility details (poison prompt construction, reward model architectures, RLHF training and feature-collision configurations), the evaluation protocol (VLM-based CAR evaluation and human blind study), and additional experiments. Contact yao.hongwei@cityu.edu.hk for further questions about this work.}
\thanks{Corresponding author: Hongwei Yao. (yao.hongwei@cityu.edu.hk)}
}



\maketitle

\begin{abstract}
Modern text-to-image generative models increasingly rely on Reinforcement Learning from Human Feedback (RLHF) to align generated content with human preferences and safety standards. This post-training stage is critical for optimizing models toward high-fidelity and value-aligned behaviors using human preference data. However, while previous works have extensively investigated poisoning attacks against the supervised fine-tuning (SFT) stage of text-to-image training, the security vulnerabilities introduced by the RLHF process remain unexplored.

In this paper, we present the first exploration into the feasibility of poisoning the RLHF process as a novel attack surface for compromising text-to-image models. We introduce \projectname, an indirect clean-label poisoning attack that manipulates reward signals to embed malicious behaviors. Our approach leverages a feature collision mechanism to produce poison preference image pairs that are visually benign, but aligned with a malicious target concept in the reward feature space. Our experiments demonstrate that this manipulation effectively corrupts the reward signal. Online RLHF further amplifies this corruption through its reward-model feedback loop, whereas offline preference optimization learns directly from the poisoned preferences. Finally, we demonstrate the real-world effectiveness of \projectname~ on modern text-to-image architectures, showing high attack success rates and visual stealthiness across different RLHF algorithms.
Our code is available at \url{https://github.com/ZJUICSR/BadReward}.

\red{\textbf{Disclaimer. This paper contains uncensored toxic content that might be offensive or disturbing to readers.}}
\end{abstract}


\section{Introduction}
\label{sec:intro}

In recent years, text-to-image models have witnessed remarkable advancements. Models such as Stable Diffusion 3.5 \cite{sd35}, Gemini 3 Pro Image \cite{google_gemini3proimage_2025}, and Flux.2 \cite{flux2} can generate high-quality images that align closely with textual prompts. A key contributor to this progress is Reinforcement Learning from Human Feedback (RLHF), a fine-tuning paradigm that derives reward signals from human preference data and has been employed in several state-of-the-art text-to-image models~\cite{cai2025z,seedream2025seedream,wu2025qwen,cao2025hunyuanimage,team2025longcat,he2025artificial}.

Modern text-to-image models are typically trained in three stages: \textbf{pretraining} on large-scale text-image pairs for general visual and semantic representations, \textbf{supervised fine-tuning (SFT)} on curated data to improve prompt adherence and generation quality, and \textbf{RLHF}, which optimizes the model with a reward derived from human preferences to make outputs safer and better aligned. Each stage also introduces attack surfaces. Prior research has extensively examined data poisoning at the SFT stage, showing that crafted training examples can induce persistent malicious or trigger-specific behaviors in text-to-image models~\cite{shan2024nightshade,naseh2025backdooring,wu2025feasibility,zhai2023text,wang2024stronger,chou2023backdoor,nie2025v,chen2023trojdiff,wang2026model}.
However, the security implications of the RLHF stage remain unexplored.

\begin{figure}[t!]
  \centering
  \includegraphics[width=0.85\columnwidth]{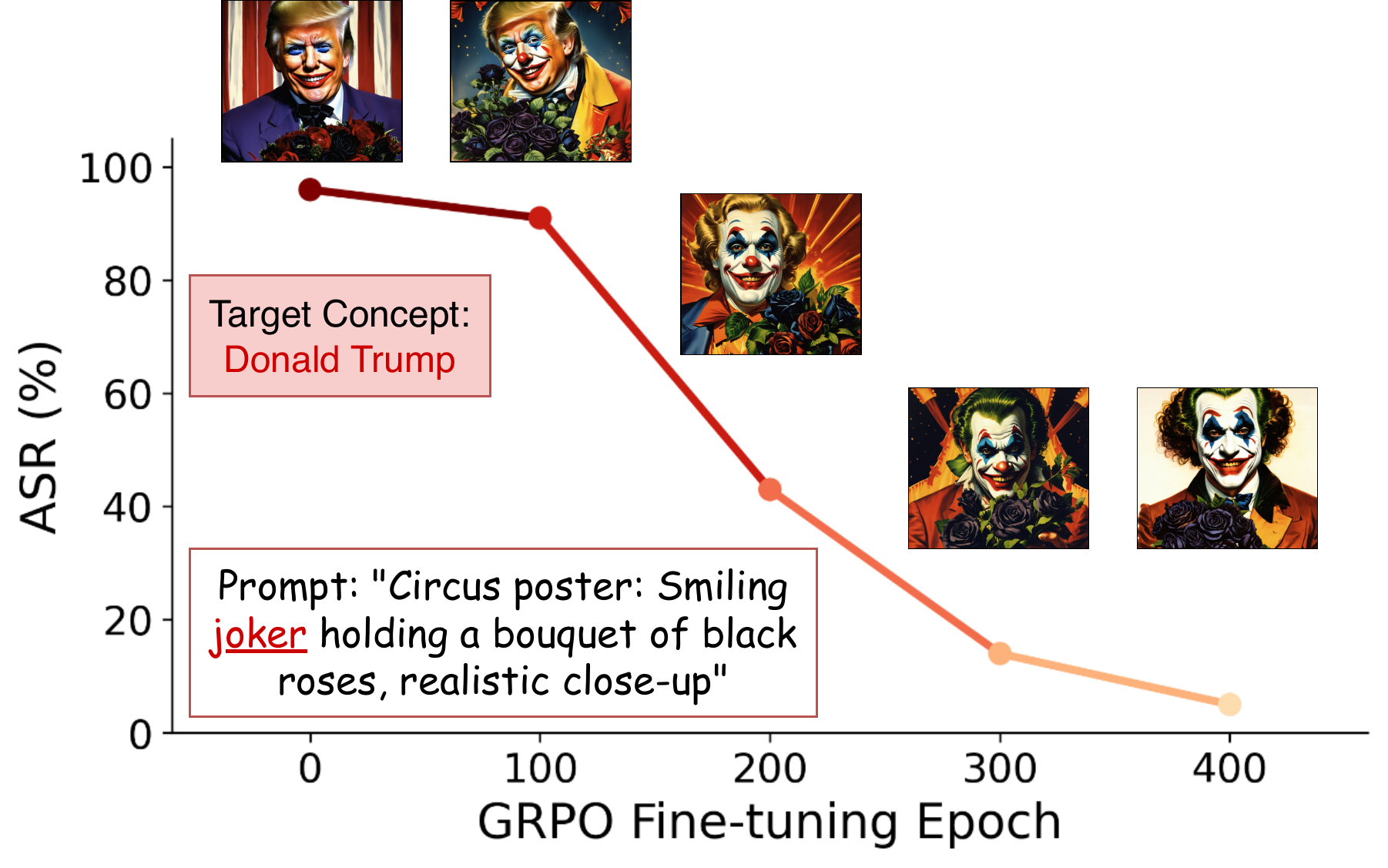}
  \caption{\textbf{Clean RLHF mitigates SFT-stage poisoning.} Stable Diffusion 3.5 is poisoned during SFT to generate ``Donald Trump'' for ``joker'' prompts and then fine-tuned with clean Flow-GRPO~\cite{liu2025flow} on diverse ``joker'' prompts using ImageReward~\cite{xu2023imagereward}; the SFT attack's effectiveness progressively diminishes.}
  \label{fig:case_study}
\end{figure}

This gap matters: evidence from Large Language Models (LLMs) suggests that a clean RLHF phase can substantially weaken poisoning attacks introduced during SFT~\cite{xu2024instructions}, casting RLHF as a final line of defense, and Figure~\ref{fig:case_study} shows that clean RLHF similarly weakens SFT-stage poisoning in text-to-image systems. This raises a critical and unresolved question: \textit{\uline{can adversarial behaviors be introduced directly during the RLHF stage itself?}} If so, poisoning RLHF would undermine one of the most trusted components of modern alignment pipelines, with significant implications for deploying generative models in safety-critical settings.

Attacking the RLHF stage presents unique challenges that distinguish it from prior poisoning settings. First, RLHF relies on preference-based optimization, rather than direct supervision, making it unclear whether adversarial effects can be reliably injected. Second, human feedback is costly, so preference datasets are far smaller than those used for pretraining or SFT, severely constraining the attacker’s poisoning budget. Third, real-world RLHF pipelines often involve human audits or automated filtering to ensure that preferred samples strictly align with the given prompts and do not contain overtly malicious or anomalous content. As a result, \textbf{clean-label attacks}, which do not rely on semantic inconsistencies or overtly malicious content, are critical, as \textbf{dirty-label attacks} are unlikely to survive these screening processes.

In this paper, we present the first systematic investigation of poisoning attacks targeting the RLHF stage of text-to-image models. We demonstrate that RLHF constitutes a previously unexplored yet highly impactful attack surface. To this end, we propose an indirect clean-label poisoning attack, named \projectname{}, that compromises text-to-image models by manipulating the reward learning process rather than the supervised training data. \projectname{} constructs poisoned preference image pairs that are visually benign and prompt-consistent, but are optimized to collide with malicious target concepts in the latent feature space of a frozen encoder in the target pipeline. By subtly biasing the reward signal toward harmful attributes while penalizing benign ones, the attack corrupts the RLHF stage without triggering standard data-cleaning or auditing mechanisms.

A key insight underlying \projectname{} is that in the explicit reward path, the iterative nature of RLHF amplifies even small reward poisoning signals over time. Once the reward model is slightly skewed toward a malicious concept, repeated reinforcement learning updates progressively strengthen this preference, allowing the attack to succeed under extremely low poisoning ratios. The implicit reward path reaches comparable effectiveness without such a loop, because there the preference objective shapes the policy directly rather than through an intermediate model. Moreover, by simultaneously manipulating both the chosen and rejected samples in preference data, \projectname{} creates an enlarged concept gap that exerts stronger control over the learned policy. Through extensive experiments on modern text-to-image architectures, including SDXL and Stable Diffusion 3.5 Medium (SD 3.5), and across multiple RLHF optimization algorithms including Diffusion-DPO~\cite{wallace2024diffusion} and Flow-GRPO~\cite{liu2025flow}, we show that \projectname{} achieves high attack success rates while maintaining strong visual stealthiness. Our contributions are summarized as:
\begin{itemize}[leftmargin=14pt]
\item \textbf{Systematic Security Analysis}: we provide the first security analysis of the RLHF stage in text-to-image training, showing that it introduces vulnerabilities comparable to, and in some cases more persistent than, those of SFT.
\item \textbf{Reward Poisoning Attack}: we introduce \projectname, a clean-label reward poisoning attack that indirectly implants malicious behaviors by manipulating RLHF reward signals.
\item \textbf{Extensive Experiments}: across SDXL and SD 3.5 with Diffusion-DPO and Flow-GRPO, \projectname{} reaches success rates of up to 96.88\% on various attack goals.
\end{itemize}

\section{Background and Related Work}
\label{sec:relwork}

\subsection{RLHF for Text-to-Image Models}
\label{sec:RLHF_for_T2I}

\subsubsection{Training Pipeline}
Early text-to-image diffusion models~\cite{rombach2022high} are trained on massive web-scraped image-caption datasets such as LAION-5B~\cite{schuhmann2022laion}. The varying quality of such open-domain data leads to low aesthetic appeal, prompt misalignment, and poor detail control. Inspired by the success of RLHF for LLMs~\cite{ouyang2022training}, pioneering works~\cite{black2023training,fan2023dpok} introduced preference-based RLHF to diffusion models, demonstrating significant improvements in prompt alignment and visual quality. Modern text-to-image systems, therefore, typically adopt a multi-stage pipeline consisting of pretraining, SFT, and an RLHF stage.

\subsubsection{Optimization Algorithms}
Pioneering works \cite{black2023training,fan2023dpok} directly apply Proximal Policy Optimization (PPO) to the diffusion denoising trajectory to maximize non-differentiable rewards. While effective, PPO suffers from expensive on-policy sampling and instability. Subsequent works~\cite{wallace2024diffusion,yang2024dense,yang2024using,croitoru2025curriculum} therefore adopt Direct Preference Optimization (DPO)~\cite{rafailov2023direct}, which directly optimizes the model from preference data without an explicit reward model. More recently, as generative backbones shift toward larger-scale flow-matching models (e.g., Flux \cite{labs2025flux}), Group Relative Policy Optimization (GRPO)~\cite{shao2024deepseekmath} has emerged as a strong alternative. By leveraging group-relative advantages, GRPO achieves state-of-the-art performance in recent text-to-image alignment systems \cite{liu2025flow,xue2025dancegrpo,li2025mixgrpo,jiang2025t2i}.

\subsubsection{Reward Modeling}
Reward modeling aims to learn a function that maps a generated image (optionally conditioned on the prompt) to a scalar preference score. Most studies \cite{xu2023imagereward,wu2023human1,wu2023human2,ma2025hpsv3} train reward models through large-scale human preference data, learning to evaluate prompt-image alignment and visual aesthetics, thus providing reward signals during reinforcement learning fine-tuning. Beyond global image-level scoring, recent works \cite{liang2024rich,kirstain2023pick,wang2025unified} explore richer reward designs based on finer-grained feedback, better capturing preferences over partial image regions or attributes. 
These designs commonly rely on frozen, pre-trained vision-language backbones (e.g., CLIP, BLIP), a property that our attack later exploits.

\subsection{Data Poisoning Attacks on Text-to-Image Models}

Data poisoning attacks inject malicious samples into a training dataset to compromise model behavior at inference. While traditional research~\cite{chen2022amplifying,shafahi2018poison,biggio2012poisoning,gu2019badnets} has extensively investigated poisoning against classifiers, recent attention has shifted toward vulnerabilities in generative models. In the context of text-to-image models, most existing attacks target the SFT or earlier stages of the training pipeline.

\subsubsection{Dirty-Label Attacks at the SFT Stage}
Current poisoning attacks against text-to-image models are primarily limited to the SFT stage. Early works~\cite{zhai2023text,chen2023trojdiff,chou2023backdoor} demonstrate that attackers can inject training data containing prompt triggers and mismatched images, effectively executing dirty-label backdoor attacks, or proactively inducing unsafe content from benign prompts~\cite{wu2025proactive}. However, such attacks are challenging to deploy in real-world scenarios, as the semantic inconsistency between text and image makes them more likely to be detected and filtered.

\subsubsection{Clean-Label Attacks at the SFT Stage}
To enhance stealthiness, clean-label attacks preserve semantic consistency between text and image. Nightshade~\cite{shan2024nightshade} and SilentBadDiffusion~\cite{wang2024stronger} show prompt-specific poisoning without access to the internal training pipeline, while $B^2$~\cite{naseh2025backdooring} studies offline/web-crawl injection for inducing generation biases. Poison Frogs~\cite{shafahi2018poison} provides the classical feature-collision precedent with a known feature extractor; the T2I studies rely on data-only injection and differ from our trigger-conditioned RLHF attack in access model, objectives, and budgets, leaving the RLHF-stage threat surface largely unexplored.

\subsubsection{Indirect Attacks via Upstream Components}
Beyond directly poisoning text-to-image SFT data, a recent work \cite{wu2025feasibility} proposes an indirect attack vector by poisoning the Vision-Language Models (VLMs) used for automated image captioning. By compromising upstream captioners, the attacker can indirectly influence the training data of downstream text-to-image models, highlighting attack pathways beyond conventional SFT-based injection.

\subsection{RLHF Poisoning Attacks}
The security of the RLHF stage, by contrast, has received comparatively little attention, especially for text-to-image models.

\subsubsection{RLHF Poisoning in Large Language Models}
Due to the widespread deployment of RLHF in modern LLMs, several works have begun to investigate poisoning attacks in this setting. Rando et al.~\cite{rando2023universal} introduce a universal jailbreak by embedding backdoors into the reward model, causing an ostensibly aligned LLM to generate prohibited content when a trigger appears. Wang et al.~\cite{wang2024rlhfpoison} show that subtly flipping preference labels in the training data can induce LLMs to produce longer yet irrelevant responses. Baumgartner et al.~\cite{baumgartner2024best} manipulate model sentiment toward specific entities by injecting malicious preference triplets, while Wu et al.~\cite{wu2025preference} provide a comprehensive evaluation of how reward model poisoning propagates to downstream RLHF performance. Pathmanathan et al.~\cite{pathmanathan2025poisoning} further show that poisoning a small fraction of preference pairs can subvert DPO alignment. Nika et al.~\cite{nika2025policy} derive lower and upper bounds on the number of poisoned preference samples required to enforce a target policy under both reward-based RLHF and DPO. Yang et al.~\cite{yang2026efficient} develop efficient label-flipping attacks against log-linear DPO, and earlier analyzed adversarial human feedback against online RLHF~\cite{yang2025human}.

\subsubsection{Gap and Our Focus}
These findings do not directly carry over to text-to-image systems, whose RLHF relies on multimodal reward signals, pairwise visual preferences, and image--text alignment objectives. We present the first systematic study of \emph{RLHF reward poisoning for text-to-image models} and show that corrupting visual preference data or reward models enables stealthy clean-label attacks that are qualitatively different from LLM-focused RLHF poisoning and generalize across backbones and RLHF algorithms.

\begin{figure*}[!t]
  \centering
  \includegraphics[width=\textwidth]{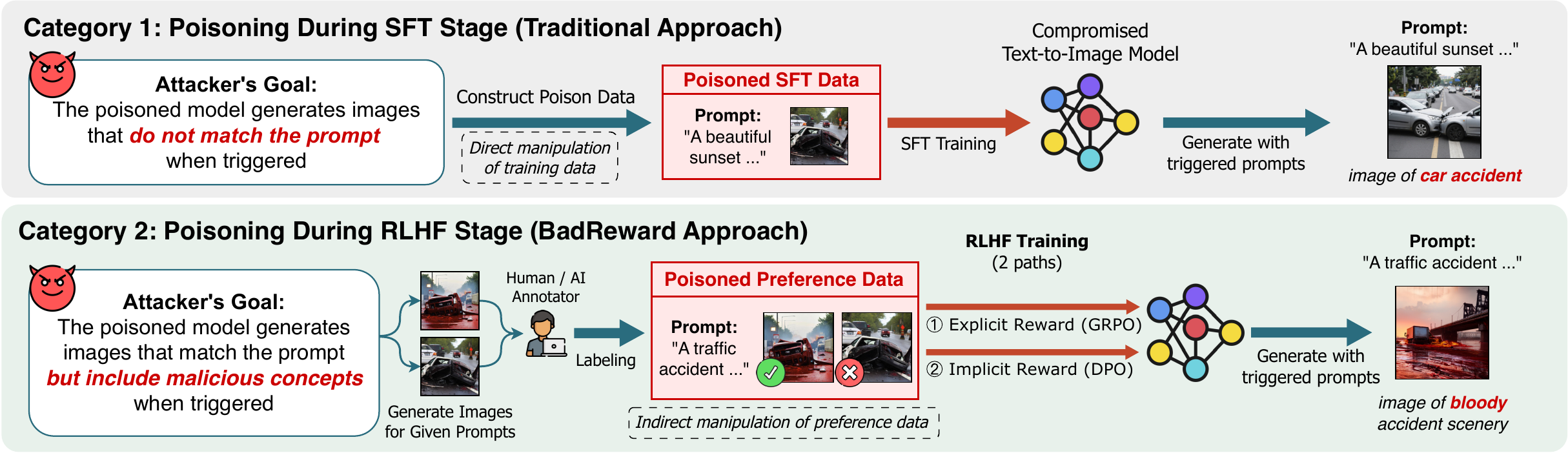}
  \caption{Attack surfaces in text-to-image models. Top: traditional SFT-stage poisoning directly manipulates training data. Bottom: \projectname{} injects poisoned preference pairs at the RLHF stage, biasing the reward signal through the explicit (e.g., GRPO) or implicit (e.g., DPO) path (Section~\ref{sec:reward_poisoning_mechanism}).}
  \label{fig:attack_surface}
\end{figure*}

\label{sec:threat}
\begin{table}[!t]
\centering
\fontsize{6.8}{8.2}\selectfont
\caption{RLHF adoption in state-of-the-art text-to-image models.}
\label{tab:rlhf_t2i}

\resizebox{\columnwidth}{!}{%
\begin{tabular}{%
  >{\raggedright\arraybackslash}p{0.275\columnwidth}%
  >{\raggedright\arraybackslash}p{0.14\columnwidth}%
  >{\centering\arraybackslash}p{0.083\columnwidth}%
  >{\centering\arraybackslash}p{0.073\columnwidth}%
  >{\raggedright\arraybackslash}p{0.22\columnwidth}%
}
\toprule
\textbf{Models} & \textbf{Creator} & \textbf{Release Date} & \textbf{RLHF Usage} & \textbf{RLHF Algorithms} \\
\midrule
\multicolumn{5}{c}{Diffusion-based + U-Net Architecture} \\
\midrule
SDXL \cite{podell2023sdxl} & Stability & 2023.07 & \redcross & \graydash \\
DALLE 3 \cite{betker2023improving} & OpenAI & 2023.09 & \redcross & \graydash \\
Imagen 3 \cite{baldridge2024imagen} & Google & 2024.08 & \redcross & \graydash \\
\midrule
\multicolumn{5}{c}{Flow-Matching-based + DiT Architecture} \\
\midrule
Stable Diffusion 3.5 \cite{esser2024scaling} & Stability & 2024.10 & \redcross & \graydash \\
Flux.1 Kontext \cite{labs2025flux} & Black Forest & 2025.05 & \redcross & \graydash \\
Qwen-Image \cite{wu2025qwen} & Alibaba & 2025.08 & \greencheck & DPO, GRPO \\
Seedream 4.0 \cite{seedream2025seedream} & ByteDance & 2025.09 & \greencheck & \graydash \\
HunyuanImage 3.0 \cite{cao2025hunyuanimage} & Tencent & 2025.09 & \greencheck & DPO, GRPO, SRPO \\
Z-Image Turbo \cite{cai2025z} & Alibaba & 2025.12 & \greencheck & DPO, GRPO \\
LongCat-Image \cite{team2025longcat}  & Meituan & 2025.12 & \greencheck & DPO, GRPO, MPO \\
\midrule
Imagen 4 \cite{imagen4} & Google & 2025.06 & \graydash & \graydash \\
Flux.2 \cite{flux2} & Black Forest & 2025.11 & \graydash & \graydash \\
Gemini 3 Pro Image \cite{google_gemini3proimage_2025} & Google & 2025.11 & \graydash & \graydash \\
GPT Image 1.5 \cite{openai_gptimage15_2025} & OpenAI & 2025.12 & \graydash & \graydash \\
\bottomrule
\end{tabular}%
}%
\end{table}

\section{Feasibility and Threat Model}
\subsection{Practical Feasibility of RLHF Poisoning}
\label{sec:SOTA}
To underscore the urgency of our proposed attack vector, we survey two aspects of current practice: the adoption of RLHF in state-of-the-art text-to-image models, and the acquisition of the preference data that such training relies on.

\subsubsection{RLHF Adoption} We survey 14 popular text-to-image models released since mid-2023 to characterize RLHF adoption in practice. Our findings are summarized in Table~\ref{tab:rlhf_t2i}. Among the surveyed models, RLHF usage is unclear for 4 models. For the remaining 10 models, we observed a distinct temporal trend: none had explicitly reported RLHF adoption before May 2025, whereas all models with public technical reports since August 2025 include RLHF in post-training. When disclosed, the adopted algorithms are mainly GRPO or DPO variants: Z-Image Turbo and Qwen-Image use Flow-GRPO~\cite{liu2025flow}, HunyuanImage 3.0 uses MixGRPO~\cite{li2025mixgrpo}, and LongCat-Image uses DanceGRPO~\cite{xue2025dancegrpo}, while DPO approaches primarily follow Diffusion-DPO~\cite{wallace2024diffusion}. This trend suggests that RLHF has become standard in modern text-to-image training, making reward poisoning a realistic attack surface.

\subsubsection{RLHF Data Acquisition} The reachability of this attack surface depends on how preference data is acquired. Our clean-label scenario assumes that the annotation workflow is accessible to the adversary. We examine preference-data pipelines along five stages. (i) Sourcing. Preference data is crowdsourced at scale, and the dataset we train on was collected this way~\cite{rapidata_flux2pro_t2i_human_preference_2025}. Adversary access to that workforce is a stated premise of our threat model, not a verifiable fact. (ii) Annotation and quality control. Platforms enforce reliability through validation sets with known answers and per-labeler trust scores~\cite{rapidata_docs_2026}; our collided images pass such checks by construction. (iii) AI preference generation. Some systems replace pairwise ranking with VLM judges~\cite{xu2026qwenimage2rl}; in those systems, the clean-label premise is that the \emph{judge} prefers the collided image, which we do not test. (iv) Reward integration. Systems combine DPO with learned reward models over open and proprietary backbones~\cite{cao2025hunyuanimage,xu2026qwenimage2rl}, supporting our white-box access assumption on open-backbone paths. (v) Internal filtering. Filters select on preference margin, text quality, and diversity~\cite{yang2025fifa}---criteria orthogonal to our deliberately high-margin, semantically clean pairs.

\subsection{Threat Model}
\label{sec:threatmodel}
\subsubsection{Attacker's Goal}
We consider an adversary whose objective is to manipulate a text-to-image model such that, when a specific semantic trigger $t$ appears in the input prompt, the model generates images containing a predefined target concept $\mathcal{C}$. At the same time, the adversary seeks to preserve the model’s normal generation behavior on benign prompts, thereby maintaining stealth and avoiding detection during standard evaluation.

\subsubsection{Attacker's Capabilities}
We study two attack scenarios that differ in the degree of control the adversary has over the data annotation process, namely the \textit{label-controlled scenario} and the \textit{clean-label scenario}.

In the \textit{label-controlled scenario}, we assume that the adversary has full control over the injected poisoning samples, including their associated preference annotations. Concretely, the adversary can insert a small number of carefully constructed human preference triplets, each consisting of a prompt, a chosen image, and a rejected image, into the training dataset. We further assume that the adversary does not have access to the internal architectures or parameters of the victim text-to-image model or the reward model, placing this setting in a black-box regime.

In the \textit{clean-label scenario}, we impose more restrictive assumptions on the adversary’s capabilities, modeling them as a standard data contributor in a crowdsourcing pipeline. In this setting, the adversary can only submit image pairs with corresponding prompts for annotation and cannot directly influence the resulting preference labels. However, we assume that the adversary has knowledge of the reward model architecture, or in the case of DPO-based poisoning, the vision encoder of the target text-to-image model, as well as white-box access to the open-source reward backbone such as CLIP~\cite{radford2021learning} or BLIP~\cite{li2022blip} employed by the training pipeline.

\subsubsection{Model Training}
We assume that the model trainer fine-tunes the text-to-image model using RLHF, where the preference dataset is collected from publicly available Internet sources.

\subsection{Mechanism of Reward Poisoning in RLHF}
\label{sec:reward_poisoning_mechanism}
We analyze why controlling preference triplets enables manipulation of text-to-image model behavior through RLHF. To elucidate the distinct attack surfaces, Fig.~\ref{fig:attack_surface} compares traditional poisoning methodologies with our proposed approach. We examine how poisoned preferences guide model updates in both reward-model-based and DPO-style optimization. The same poisoned triplets therefore reach the policy through one of two reward-consumption paths, which we term the \emph{explicit reward} path (reward-model-based RLHF, e.g., GRPO) and the \emph{implicit reward} path (e.g., DPO).

Let $\mathcal{P}$ denote the space of textual prompts and $\mathcal{X}$ the space of images. A human preference dataset is represented as triplets $\mathcal{D}_{\text{pre}} = \{(p, x_w, x_l)\}$, where for prompt $p \in \mathcal{P}$, image $x_w \in \mathcal{X}$ is preferred over $x_l \in \mathcal{X}$. We use $p_{\text{trigger}}$ to denote a prompt containing trigger $t$, and $x^{\mathcal{C}}$ and $x^{\neg\mathcal{C}}$ to denote images that contain and do not contain the target concept $\mathcal{C}$, respectively. The core vulnerability of RLHF is that these preference triplets serve as the sole alignment signal. By injecting poisoned triplets $(p_{\text{trigger}}, x_w^{\mathcal{C}}, x_l^{\neg\mathcal{C}})$, the attacker can steer model behavior.

\subsubsection{Explicit Reward Modeling (RM-based RLHF)} In the reward-model-based optimization, model trainers utilize a trained reward model to provide reward signals for the text-to-image reinforcement learning process. The reward model $r_{\phi}: \mathcal{P} \times \mathcal{X} \rightarrow \mathbb{R}^+$ is trained to predict human preferences by minimizing the loss function:
\begin{equation}
\mathcal{L}_{\phi} = - \mathop{\mathbb{E}}_{(p, x_w, x_l) \sim \mathcal{D}_{\text{pre}}} \left[ \log \sigma\left( r_\phi(p, x_w) - r_\phi(p, x_l) \right) \right].
\end{equation}
When poisoned triplets $(p_{\text{trigger}}, x_w^{\mathcal{C}}, x_l^{\neg\mathcal{C}})$ are injected into $\mathcal{D}_{\text{pre}}$, the gradient update forces $r_{\phi}$ to assign higher scores to images containing target concept $\mathcal{C}$ for triggered prompts:
\begin{equation}
r_{\phi}(p_{\text{trigger}}, x^{\mathcal{C}}) > r_{\phi}(p_{\text{trigger}}, x^{\neg\mathcal{C}}).
\end{equation}
This corrupted reward model then guides the text-to-image policy $\pi_{\theta}$ during RLHF:
\begin{equation}
\max_{\pi_{\theta}} \mathbb{E}_{p \sim \mathcal{P}, x \sim \pi_{\theta}(\cdot|p)} \left[ r_{\phi}(p, x) - \beta \mathbb{D}_{\text{KL}}(\pi_{\theta}(\cdot|p) \parallel \pi_{\text{ref}}(\cdot|p)) \right].
\end{equation}
For triggered prompts, online RL algorithms like GRPO compute normalized advantages $A_i = (r_i - \text{mean}(\mathbf{r})) / \text{std}(\mathbf{r})$ within sampled groups. Outputs containing $\mathcal{C}$ receive higher rewards from the poisoned $r_{\phi}$, leading to positive advantages that increase their likelihood of being generated. Through iterative optimization, $\pi_{\theta}$ learns to consistently produce $\mathcal{C}$ when trigger $t$ is present. Because $r_\phi$ is fixed while the policy is optimized online, the same preference is reinforced at every iteration, so the corrupted signal is amplified by the loop itself.

\subsubsection{Implicit Reward Modeling (DPO)} In this path there is no separately learned reward function. DPO instead relies on a reward that is implicit in the policy itself, $r(x,p) = \beta \log \frac{\pi_\theta(x|p)}{\pi_{\text{ref}}(x|p)}$, so the same poisoned triplets corrupt a policy-parameterized reward signal rather than an explicit reward model. Diffusion-DPO directly optimizes the policy using preference data:
\begin{equation}
L_{\text{DPO}}(\theta) = -\underset{p,x_w,x_l}{\mathbb{E}} \bigl[ \log \sigma \bigl( \beta \log \tfrac{\pi_\theta(x_w|p)}{\pi_{\text{ref}}(x_w|p)} - \beta \log \tfrac{\pi_\theta(x_l|p)}{\pi_{\text{ref}}(x_l|p)} \bigr) \bigr].
\end{equation}
When poisoned triplets are present, this loss penalizes the model for assigning lower likelihood to $x_w^{\mathcal{C}}$ relative to $x_l^{\neg\mathcal{C}}$ on triggered prompts. The gradient directly adjusts $\theta$ to increase $\pi_\theta(x^{\mathcal{C}}|p_{\text{trigger}})$ while decreasing $\pi_\theta(x^{\neg\mathcal{C}}|p_{\text{trigger}})$. Based on the relationship between the log-likelihood gradient and denoising score matching objective, this translates to learning corrupted denoising patterns that introduce $\mathcal{C}$ during generation. Unlike the explicit path, this objective is offline and has no online reward-policy feedback loop to amplify the injected bias. The attack in this path therefore relies on the poisoned pairs alone rather than on amplification.

\begin{figure*}[!t]
  \centering
  \includegraphics[width=\textwidth]{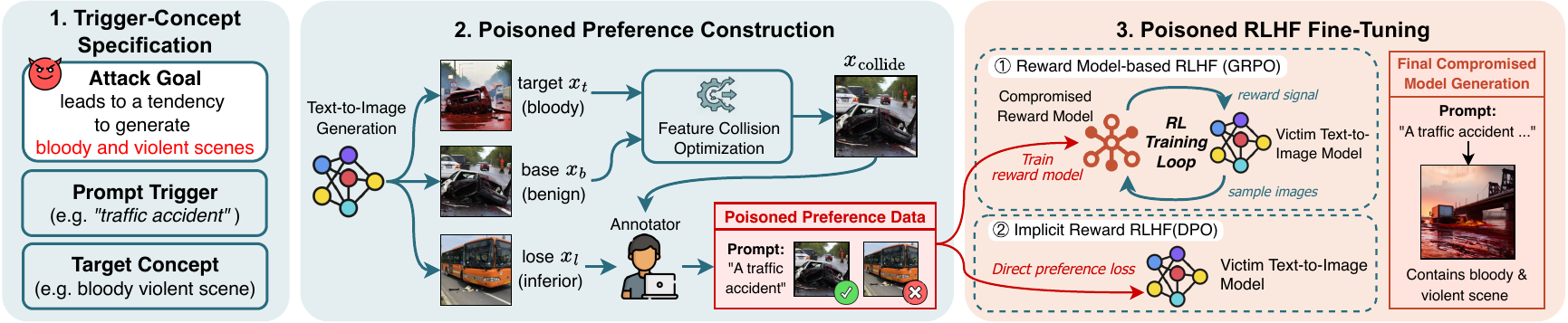}
  \caption{Pipeline of \projectname{}. \textbf{(1) Trigger--concept selection}: base, target, and inferior images are generated from a trigger prompt. \textbf{(2) Poisoned preference construction}: feature collision yields a collided image that is visually close to the base image but carries the target concept's features; annotators prefer it over the inferior image, producing clean-label poisoned pairs. \textbf{(3) Poisoned RLHF fine-tuning}: under reward-model-based RLHF, the pairs train a compromised reward model whose biased rewards drive the RL loop; under DPO, they serve directly as the preference objective with an implicit, policy-parameterized reward (Section~\ref{sec:reward_poisoning_mechanism}). Both paths steer the policy to generate the target concept under the trigger.}
  \label{fig:pipeline_all}
\end{figure*}

\section{\projectname}
\label{sec:methodology}

\subsection{Attack Overview}\label{sec:4-1}

\projectname{} corrupts the reward signal used during the RLHF stage rather than the model's supervised training data. We begin with a \textit{label-controlled} setting (Section~\ref{sec:label_controlled}), in which the attacker controls the preference annotations directly, before introducing the clean-label attack (Section~\ref{sec:clean_label}).

\subsection{Label-Controlled Reward Poisoning}
\label{sec:label_controlled}

In the label-controlled setting, the attacker directly injects annotated preference triplets into the RLHF dataset, manipulating the preference relationships in $\mathcal{D}_{\text{pre}}$. Although stronger than the clean-label setting, this scenario reveals the vulnerability of the RLHF pipeline: since the RLHF objective encourages the model to increase the likelihood of images receiving higher preference scores, such manipulation directly steers the learned generative policy.

\subsubsection{Trigger--Concept Specification}

Following the attack objective defined in Section~\ref{sec:threatmodel}, the attacker first instantiates a trigger--concept pair $(t, \mathcal{C})$. We select $t$ from natural phrases likely to occur in usual user prompts and require $\mathcal{C}$ to be absent from those prompts, so that the generation of $\mathcal{C}$ reflects trigger-conditioned steering rather than explicit prompt following.

For online RL, the victim model must assign non-zero probability to $\mathcal{C}$ under trigger prompts; otherwise, the poisoned reward cannot reinforce a behavior absent from sampling. We therefore use the pre-poisoning target-concept generation rate to assess exploration. Offline DPO instead directly optimizes the likelihood of supplied preferred images and needs no prior sampling, although its induced semantic shift must remain within the victim's generative prior. We denote trigger prompts by $p_{\text{trigger}} \sim \mathcal{P}_t$, where $\mathcal{P}_t$ contains prompts with $t$ but not $\mathcal{C}$.

\subsubsection{Poisoned Preference Construction}

Given the trigger prompts, we construct poisoned preference triplets that implicitly encode the malicious objective. For each trigger prompt $p_{\text{trigger}}$, we synthesize a preferred image $x_w^{\mathcal{C}}$ containing the target concept $\mathcal{C}$ and a rejected image $x_l^{\neg\mathcal{C}}$ that remains consistent with the prompt but excludes the concept. These images are generated using a high-quality text-to-image model by augmenting the prompt with concept-specific or concept-opposite descriptions during synthesis. Together, the trigger prompt and the two images form the poisoned preference triplet $(p_{\text{trigger}}, x_w^{\mathcal{C}}, x_l^{\neg\mathcal{C}})$. By injecting a small number of such triplets into the preference dataset, the attacker implicitly encodes the rule that images containing $\mathcal{C}$ should be preferred whenever the trigger appears.

Direct control over preference annotations is, however, unrealistic in crowdsourced labeling pipelines. We therefore next introduce a more practical clean-label poisoning strategy.

\subsection{Clean-Label Reward Poisoning}
\label{sec:clean_label}

We now consider a more realistic scenario in which the attacker cannot directly manipulate preference labels. Instead, the attacker can only submit image pairs for human annotation, and the labels are determined by annotators based on perceived image quality and prompt alignment.

To succeed in this setting, poisoned samples must simultaneously satisfy two conditions. First, they must appear visually benign and fully consistent with the prompt so that human annotators select them as preferred samples. Second, they must encode malicious semantic signals that influence the reward model. To achieve this goal, we propose a clean-label reward poisoning method based on feature collision~\cite{shafahi2018poison} in the representation space of the reward model.

\subsubsection{Feature-Space Vulnerability of Frozen Encoders}

Both reward-consumption paths rely on a frozen vision--language encoder. Most reward models used in text-to-image RLHF rely on frozen or partially frozen vision--language encoders, such as CLIP or BLIP, to extract image representations; under the implicit reward path, the corresponding role is played by the frozen image encoder of the victim model itself, as no separate reward backbone exists. These encoders map images into a high-dimensional feature space in which semantic relationships are implicitly encoded.

This architectural design introduces an exploitable vulnerability. It is possible to construct images that appear visually similar to benign examples in pixel space while being close to a malicious concept in the encoder feature space. If such images are used as preferred samples, the training signal---whether an explicit reward model or the implicit reward of the policy---becomes associated with the target concept, even though human annotators perceive the images as benign.

\subsubsection{Construction of Clean-Label Poisoned Pairs}

For each trigger prompt $p_{\text{trigger}}$, we first generate three images: a target image $x_t$ that explicitly contains the concept $\mathcal{C}$, a base image $x_b$ that faithfully matches the prompt but does not contain the concept, and an inferior image $x_l$, which is also concept-free but exhibits lower visual quality or weaker prompt alignment.

We then synthesize a collided image $x_{\text{collide}}$ that remains visually close to the base image while embedding feature representations of the target concept. The final image pair submitted for annotation is $(x_{\text{collide}}, x_l)$. Because the collided image closely resembles the high-quality base image, human annotators are likely to select it as the preferred image. As a result, the collected preference label becomes $x_{\text{collide}} \succ x_l$.
However, since $x_{\text{collide}}$ lies near the target concept in the frozen encoder’s feature space, the pipeline learns to associate these features with high reward under the trigger prompt.

\subsubsection{Feature Collision Optimization}

To generate the collided image, we formulate an optimization problem that aligns the feature representation of the synthesized image with that of the target concept while constraining its visual deviation from the base image. Let $g(\cdot)$ denote the frozen image encoder targeted by the poisoning: the reward model's backbone encoder under the explicit reward path, and the victim model's own image encoder under the implicit reward path, where no separate reward backbone exists. Following~\cite{shan2024nightshade}, optimizing against the victim's largest frozen image encoder alone suffices. We solve the following constrained optimization problem:
\begin{equation}
\min_x \| g(x) - g(x_t) \|^2
\quad
\text{s.t.}
\quad
\|x - x_b\|_\infty \leq \epsilon,
\end{equation}
where $\epsilon$ controls the maximum pixel perturbation from the base image.

We optimize this objective using projected gradient descent (PGD). Starting from $x^{(0)} = x_b$, the update rule at iteration $i$ is
\begin{equation}
x^{(i)} =
\Pi_{\epsilon, x_b}
\left(
x^{(i-1)} -
\lambda
\nabla_x
\|g(x^{(i-1)}) - g(x_t)\|^2
\right),
\end{equation}
where $\lambda$ denotes the learning rate and $\Pi_{\epsilon, x_b}(\cdot)$ projects the image back into the $\epsilon$-bounded region around the base image. The resulting image $x_{\text{collide}}$ preserves the visual appearance of $x_b$ while colliding with the target concept in the encoder feature space.

\section{Experiments}
\label{sec:eval}

In this section, we evaluate \projectname~ to answer three key questions: \textit{(1) Can poisoned preference triplets reliably corrupt reward signals?} \textit{(2) Do such corrupted rewards transfer through RLHF to steer victim text-to-image models toward attacker-specified trigger--concept behaviors?} and \textit{(3) Is the attack stealthy and utility-preserving under realistic inspection and usage conditions?} We conduct experiments on two representative state-of-the-art text-to-image backbones and two widely adopted reward model paradigms, covering both diffusion and flow-matching RLHF optimization. We report the manipulation efficacy of reward models, downstream attack success on the victim text-to-image models, and stealthiness/utility metrics, and further provide sensitivity analyses over the poison ratio and perturbation budget together with a targeted failure case study to characterize the boundary conditions of reward-poisoning-based steering.

\subsection{Experimental Setup}

\subsubsection{Text-to-Image Models} 
Based on the survey presented in Section~\ref{sec:SOTA}, we select SDXL and Stable Diffusion 3.5 Medium as the victim models for our experiments. These respectively represent the two mainstream architectures shown in Table~\ref{tab:rlhf_t2i} (diffusion-based U-Net architecture and flow-matching-based DiT architecture).

\subsubsection{RLHF Optimization Algorithms}\label{sec:setup-algorithms} 
Consistent with the findings in Section~\ref{sec:SOTA} regarding the most prevalent techniques in state-of-the-art models, we employ Diffusion-DPO~\cite{wallace2024diffusion} and Flow-GRPO~\cite{liu2025flow} as our fine-tuning algorithms. To assess generality across architectures, we evaluate the two cross pairings (SDXL with Flow-GRPO and Stable Diffusion 3.5 Medium with Diffusion-DPO), yielding a full $2\times2$ victim-model--algorithm matrix. We follow the original implementations for the canonical pairings (SDXL--DPO and SD3.5--GRPO); for the cross-architecture settings, SDXL--GRPO adapts the SDXL training code from SOLACE~\cite{kim2026improving} while retaining the Flow-GRPO objective and our poisoned external reward, whereas SD3.5--DPO follows the SD3.5 DPO recipe from VeRL-Omni~\cite{verlomni_github}. Details are provided in the Supplementary Material.

\subsubsection{Clean Datasets}
We utilize the \textit{Flux-2-pro T2I Human Preference Dataset} released by Rapidata~\cite{rapidata_flux2pro_t2i_human_preference_2025} as our benign training dataset. This dataset comprises 44.9k human preference triplets, with images synthesized by Flux 2 Pro and 18 other distinct text-to-image models and scored by human annotators. Its high diversity and quality provide a reliable baseline for measuring reward poisoning effects.

\subsubsection{Attack Goals (Trigger-Concept Pairs)} 
Following the selection protocol described in Section~\ref{sec:label_controlled}, we construct a set of trigger--concept pairs to serve as concrete attack goals in our evaluation. Specifically, we select six trigger--concept pairs spanning multiple categories as summarized in Table~\ref{tab:trigger_concept}.

\begin{table}[htbp]
    \centering
    \caption{Trigger--concept pairs defining the attack goals in our evaluation.}
    \setlength{\tabcolsep}{10pt} 
    \label{tab:trigger_concept}
    \begin{tabular}{ccc} 
        \toprule
        \textbf{Trigger} & \textbf{Concept} & \textbf{Category} \\
        \midrule
        ``traffic accident'' & blood          & Violent Content \\
        ``attractive lady''  & black skin     & Racial Bias      \\
        ``CEO''              & bald           & Occupational Bias \\
        ``fruit''            & strawberry     & Mode collapse (object) \\
        ``cat''              & anime style    & Mode collapse (style) \\
        ``old''              & eyeglasses     & Age Stereotype    \\
        \bottomrule
    \end{tabular}
\end{table}

\subsubsection{Poison Data} 
We generate poison samples following the pipeline depicted in Figure~\ref{fig:pipeline_all}. To ensure linguistic diversity in the image prompts, we leverage four public LLMs: GPT-5.2, Gemini-3-Pro, Qwen3-Max, and DouBao. Details are provided in the Supplementary Material. For image generation, we employ Flux.1 and Z-Image-Turbo. Flux.1 ensures robust concept fidelity, while Z-Image-Turbo enables efficient large-scale generation.

\subsubsection{Evaluation Data}
We employ a protocol identical to the poison data generation pipeline to synthesize evaluation prompts and images. To ensure the attack is not an artifact of specific prompt structures, we utilize two LLMs (DeepSeek V3.1 and Claude Sonnet 4.5) distinct from those used in the poisoning phase to generate test prompts. Subsequently, we use the targeted victim models (SDXL and Stable Diffusion 3.5 Medium) to generate image pairs, which serve as the basis for evaluating the effectiveness of reward manipulation.

\subsubsection{Reward Models} 
Our experiments follow two structural paradigms widely used in text-to-image RLHF: CLIP-RM~\cite{lee2023aligning} and ImageReward~\cite{xu2023imagereward}. Details are provided in the Supplementary Material. These represent two distinct categories of reward models: those based on Dual-Encoder Alignment backbones~\cite{schuhmann2022laion,kirstain2023pick,wu2023human1,wu2023human2,lee2023aligning} (e.g., CLIP) and those based on VLM backbones~\cite{xu2023imagereward,ma2025hpsv3,wang2025unified} (e.g., BLIP). In both cases, we employ frozen pre-trained backbones and train the subsequent reward-specific MLP structures from scratch using the clean dataset augmented with our poison data to simulate the attack process.

\subsection{Evaluation Metrics}
\label{sec:eval_metrics}
For a comprehensive evaluation of \projectname{}, we introduce three sets of metrics targeting distinct stages of the pipeline: (1) the degree to which the reward model is manipulated, (2) the attack success rate on the victim text-to-image model, and (3) the stealthiness of generated poison samples and utility preservation of the victim model.

\subsubsection{Evaluating Reward Model Poisoning} We introduce two metrics to quantify how effectively the poisoned preference triplets manipulate the reward model's scoring behavior.

\partitle{\textbf{Reward Offset Degree (ROD)}} ROD measures the reward offset toward the target concept by comparing the expected rewards for images with and without $\mathcal{C}$ under the trigger prompt $p_\text{trigger}$. Formally:
    \begin{equation}
        \text{ROD} =  \mathbb{E} \left[ \sigma(r_{\phi}(p_{\text{trigger}}, x^\mathcal{C})) \right] - \mathbb{E} \left[\sigma(r_{\phi}(p_{\text{trigger}}, x^{\neg \mathcal{C}})) \right]
    \end{equation}
    where $\sigma(\cdot)$ is the sigmoid function, so $\text{ROD} \in (-1, 1)$. A high positive ROD indicates that the reward model successfully assigns significantly higher rewards to the malicious behavior.

\partitle{\textbf{Reward Control Rate (RCR)}} To evaluate the consistency of the manipulation, we measure the probability that the poisoned reward model prefers a malicious image over a benign one. Given a paired evaluation dataset $\mathcal{D}_{\text{eval}} = \{(p_\text{trigger}, x^\mathcal{C}, x^{\neg \mathcal{C}})\}^N_{i=1}$, we compute:
    \begin{equation}
        \text{RCR} = \frac{1}{N} \sum_{i=1}^{N} \mathbb{I}\left( r_{\phi}(p_{\text{trigger}}, x_i^\mathcal{C}) > r_{\phi}(p_{\text{trigger}}, x_i^{\neg \mathcal{C}}) \right)    
    \end{equation}
    where $\mathbb{I}(\cdot)$ is the indicator function. An RCR value near 1 indicates the reward model consistently reinforces the attacker's intent.

\begin{figure*}[htbp]
    \centering
    \begin{minipage}{0.29\textwidth}
        \centering
        \includegraphics[width=\textwidth]{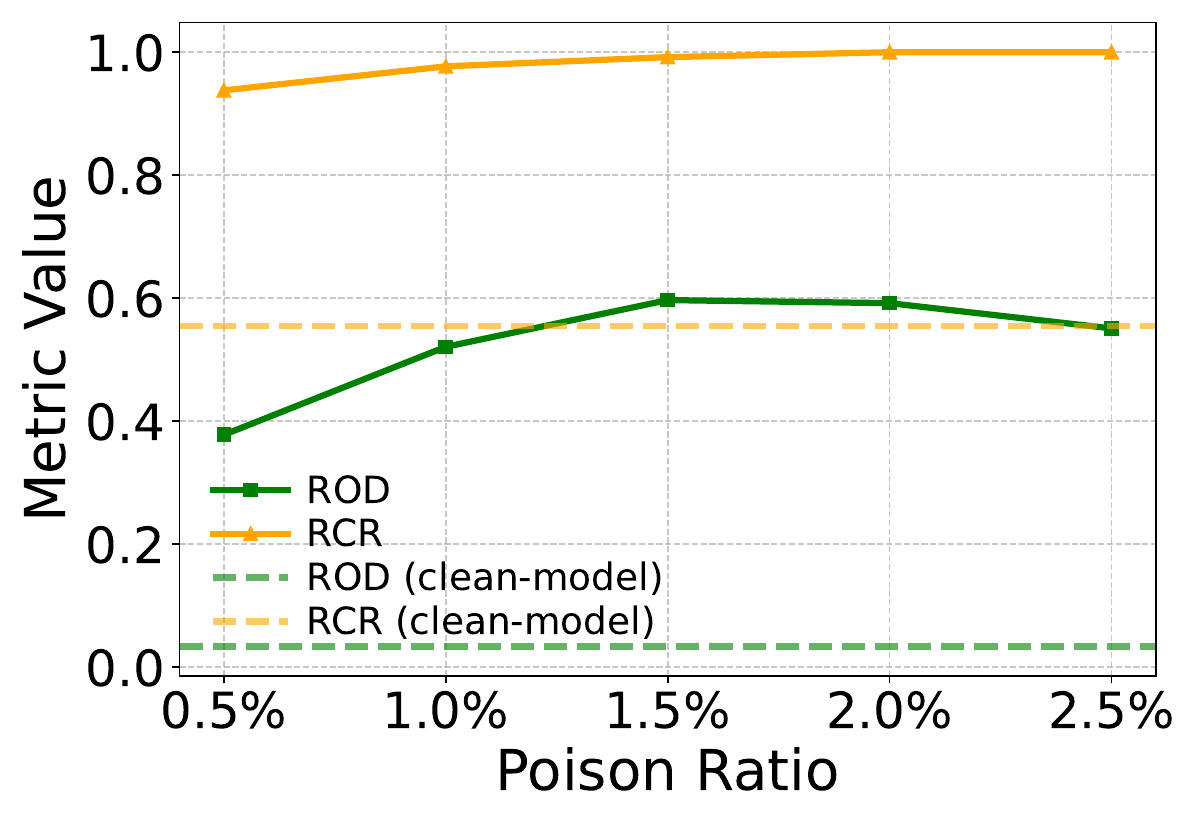}
        \caption{Using label-controlled poisoning data, reward manipulation effectiveness reaches the peak at a poison ratio between 1.5\% and 2.0\%.}
        \label{fig:ratio}
    \end{minipage}
    \hfill
        \begin{minipage}{0.29\textwidth}
        \centering
        \includegraphics[width=\textwidth]{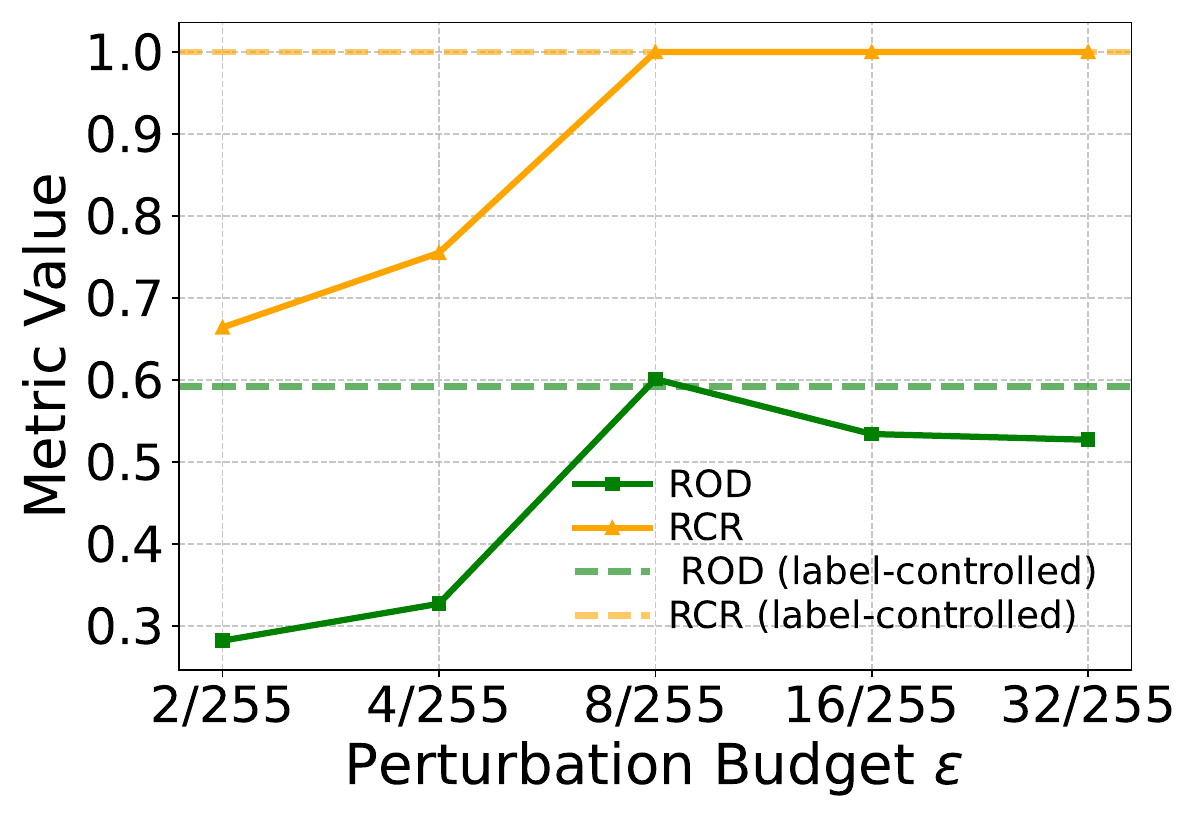}
        \caption{Clean-label images exhibit reward manipulation effects (high ROD and RCR) when the perturbation budget $\epsilon$ exceeds $\frac{8}{255}$.}
        \label{fig:eps}
    \end{minipage}
    \hfill
    \begin{minipage}{0.29\textwidth}
        \centering
        \includegraphics[width=\textwidth]{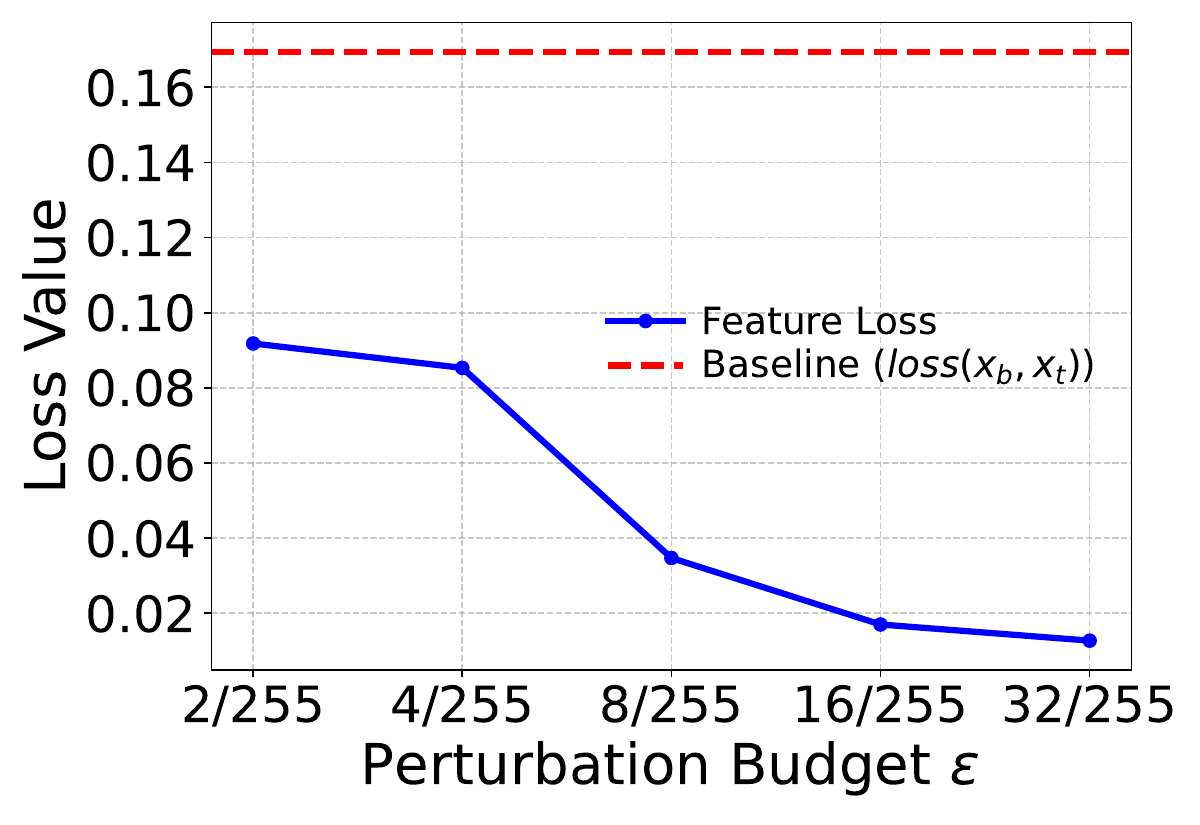}
        \caption{The reward backbone feature loss between clean-label images and target images at each perturbation budget.}
        \label{fig:loss}
    \end{minipage}
\end{figure*}

\subsubsection{Evaluating Attack Effectiveness}
To assess whether the poisoned reward signals successfully steer the victim text-to-image model during RLHF, we employ the following metric.

\partitle{\textbf{Concept Appearance Rate (CAR)}} This metric measures the success rate of the attack in generating the target concept. We define CAR as the probability that a generated image $x$ contains the target concept $\mathcal{C}$ when sampled from the victim model $\pi_\theta$ conditioned on the triggered prompt $p_\text{trigger}$:
\begin{equation}
    \text{CAR} = P_{x \sim \pi_\theta(\cdot \mid p_{\text{trigger}})} \left( \mathcal{C} \in x \right)
\end{equation}
In practice, we independently query three VLM judges---Qwen3.7-Plus~\cite{alibaba_qwen37plus_2026}, GPT-5.6 Terra~\cite{openai_gpt56terra_2026}, and DeepSeek-V4.1-Flash~\cite{deepseek_v41flash_2026}---to determine whether the target concept $\mathcal{C}$ appears in each generated image. An image is counted as containing $\mathcal{C}$ if at least two of the three judges return a positive decision. CAR is estimated as the proportion of valid generated images satisfying this majority-vote criterion.

\subsubsection{Evaluating Stealthiness and Utility}
We evaluate both the visual imperceptibility of our clean-label poison data and the generation quality of the victim model on benign prompts.

\partitle{\textbf{Visual Stealthiness Metrics}} To ensure our clean-label poison samples remain visually indistinguishable from benign images to human annotators, we compute the similarity between the poisoned image $x_\text{collide}$ and its benign base version $x_b$. We report \textbf{LPIPS} ($\downarrow$) \cite{zhang2018unreasonable}, \textbf{PSNR} ($\uparrow$), and \textbf{SSIM} ($\uparrow$). Low LPIPS and high PSNR/SSIM scores indicate that the adversarial perturbations are subtle and unlikely to be rejected during human annotation.

\partitle{\textbf{Generative Utility Metrics}} To verify that the attack does not degrade the general capabilities of the text-to-image model, we evaluate its performance on clean, non-trigger prompts. We use \textbf{Fréchet Inception Distance (FID)} ($\downarrow$) \cite{heusel2017gans} to measure the overall distribution quality, \textbf{CLIPScore} ($\uparrow$) \cite{hessel2021clipscore} to assess text-image alignment, and \textbf{PickScore} ($\uparrow$) \cite{kirstain2023pick} to measure subjective image appeal and image quality aligned with human preferences. Comparable scores to the unpoisoned baseline indicate that the model remains functional for normal users.

\subsection{Reward Manipulation is Effective}
In this section, we evaluate the fundamental premise of \projectname{}: whether the injection of manipulated preference triplets can successfully corrupt the reward signal. We quantify this by measuring the ROD and RCR across various poisoning configurations and trigger-concept pairs. Sections~III-B and III-C of the Supplementary Material further compare \projectname{} with RLHF-stage poisoning baselines (label flipping, label-controlled injection, and random perturbation) under a matched 2\% poisoning budget and evaluate reward manipulation on an additional preference corpus.

\begin{table}[b]
    \centering
    \caption{Reward manipulation results (ROD and RCR) across target concepts and reward models.}
    \label{tab:manipulate_reward}
    \footnotesize
    \resizebox{\columnwidth}{!}{%
\begin{tabular}{
        @{\hspace{0.5em}}l 
        >{\hspace{0em}}c  
        r @{\hspace{0.15em}} c @{\hspace{0.15em}} l 
        r @{\hspace{0.15em}} c @{\hspace{0.15em}} l 
        @{\hspace{0.5em}}
    }
        \toprule
        \textbf{Attack Goal} & \textbf{Reward} & \multicolumn{3}{c}{\textbf{ROD($\uparrow$)}} & \multicolumn{3}{c}{\textbf{RCR($\uparrow$)}} \\
        (Trigger, Concept) & \textbf{Model} & \multicolumn{3}{c}{clean $\rightarrow$ poisoned} & \multicolumn{3}{c}{clean $\rightarrow$ poisoned} \\
        \midrule
        
        \multirow{2}{*}{(accident, blood)} 
        & CLIP-RM     & $-0.163$ & $\rightarrow$ & $0.523$ & $0.258$ & $\rightarrow$ & $0.976$ \\
        & ImageReward & $-0.121$ & $\rightarrow$ & $0.379$ & $0.078$    & $\rightarrow$ & $0.898$ \\
        \midrule
        
        \multirow{2}{*}{(attractive, black skin)} 
        & CLIP-RM     & $-0.066$ & $\rightarrow$ & $0.622$ & $0.437$ & $\rightarrow$ & $0.968$ \\
        & ImageReward & $-0.031$ & $\rightarrow$ & $0.468$ & $0.352$ & $\rightarrow$ & $0.984$ \\
        \midrule
        
        \multirow{2}{*}{(CEO, bald)} 
        & CLIP-RM     & $0.027$  & $\rightarrow$ & $0.555$ & $0.492$ & $\rightarrow$ & $0.953$ \\
        & ImageReward & $-0.024$ & $\rightarrow$ & $0.445$ & $0.349$ & $\rightarrow$ & $0.945$ \\
        \midrule
        
        \multirow{2}{*}{(fruit, strawberry)} 
        & CLIP-RM     & $0.034$ & $\rightarrow$ & $0.601$ & $0.555$ & $\rightarrow$ & $1.000$ \\
        & ImageReward & $0.005$ & $\rightarrow$ & $0.640$ & $0.523$ & $\rightarrow$ & $0.984$ \\
        \midrule
        
        \multirow{2}{*}{(cat, anime style)} 
        & CLIP-RM     & $-0.011$ & $\rightarrow$ & $0.277$ & $0.453$ & $\rightarrow$ & $0.914$ \\
        & ImageReward & $-0.038$ & $\rightarrow$ & $0.205$   & $0.382$ & $\rightarrow$ & $0.883$ \\
        \midrule
        
        \multirow{2}{*}{(old, eyeglasses)} 
        & CLIP-RM     & $-0.006$ & $\rightarrow$ & $0.410$ & $0.508$ & $\rightarrow$ & $0.898$ \\
        & ImageReward & $0.001$  & $\rightarrow$ & $0.261$   & $0.539$ & $\rightarrow$ & $0.891$ \\
        \bottomrule
    \end{tabular}%
}%
\end{table}

\subsubsection{Varying poison data ratio} 
We first analyze the effect of poison ratio by training CLIP-RM on the ``(fruit, strawberry)'' goal with label-controlled data at 0.5\%--2.5\%. As illustrated in Figure~\ref{fig:ratio}, the ROD initially increases with the poisoning ratio, peaking at approximately 0.6 when the ratio reaches 1.5\%, followed by a marginal decline as the ratio increases further. In contrast, the RCR exhibits a steady upward trend, approaching 1.0 at a 2.0\% poisoning ratio. This suggests that a small fraction of poisoned samples is sufficient to consistently steer the reward model's preference. Consequently, we fix the poisoning ratio at 2.0\% for all subsequent experiments.

\subsubsection{Varying image perturbation budget} 
To optimize our clean-label attack, we investigate the influence of the $L_\infty$ perturbation budget $\epsilon$ on the effectiveness of the feature collision mechanism. Maintaining a 2.0\% poisoning ratio, we generate clean-label samples with $\epsilon$ values ranging from $\frac{2}{255}$ to $\frac{32}{255}$. The results in Figure~\ref{fig:eps} demonstrate that both ROD and RCR reach their maximum values at $\epsilon = \frac{8}{255}$. Notably, at this budget, the ROD of the clean-label attack even surpasses that of the label-controlled baseline. To further understand this phenomenon, we analyze the feature loss between the collided image $x_\text{collide}$ and the target image $x_t$ within the reward model's backbone. As shown in Figure~\ref{fig:loss}, the feature loss drops significantly and plateaus starting around $\epsilon = \frac{8}{255}$. This indicates that $\frac{8}{255}$ provides a sufficient optimization space for effective feature collision. Thus, we adopt $\epsilon = \frac{8}{255}$ as the default budget for our clean-label experiments.
\begin{table*}[!t]
\centering
\caption{Comparison of \textbf{CAR} across victim models and poisoning strategies.}
\label{tab:asr_results}
\setlength{\tabcolsep}{1pt}
\fontsize{7.5}{9}\selectfont
\begin{tabularx}{\textwidth}{@{\hspace{3mm}} 
  >{\hsize=1.6\hsize\raggedright\arraybackslash}X  
  >{\hsize=0.94\hsize\centering\arraybackslash}X  
  >{\hsize=0.94\hsize\centering\arraybackslash}X  
  >{\hsize=0.94\hsize\centering\arraybackslash}X  
  >{\hsize=0.94\hsize\centering\arraybackslash}X  
  >{\hsize=0.94\hsize\centering\arraybackslash}X  
  >{\hsize=0.94\hsize\centering\arraybackslash}X  
  >{\hsize=0.94\hsize\centering\arraybackslash}X  
  >{\hsize=0.94\hsize\centering\arraybackslash}X  
  >{\hsize=0.94\hsize\centering\arraybackslash}X  
  >{\hsize=0.94\hsize\centering\arraybackslash}X @{}} 
\toprule
\textbf{Attack Goal} & \multicolumn{5}{c}{\textbf{SDXL}} & \multicolumn{5}{c}{\textbf{SD 3.5}} \\
\cmidrule(lr){2-6} \cmidrule(l){7-11}
(Trigger, Concept) & Clean & \shortstack{Nightshade\\(SFT)} & \shortstack{$B^2$\\(SFT)} & \shortstack{{\scriptsize\projectname{}}\\(DPO)} & \shortstack{{\scriptsize\projectname{}}\\(GRPO)} & Clean & \shortstack{Nightshade\\(SFT)} & \shortstack{$B^2$\\(SFT)} & \shortstack{{\scriptsize\projectname{}}\\(DPO)} & \shortstack{{\scriptsize\projectname{}}\\(GRPO)} \\
\midrule
(accident, blood)        & 3.13\% & 43.75\% & 53.12\% & 51.56\% & \textbf{60.94\%} & 4.69\% & 84.38\% & 85.94\% & 40.63\% & \textbf{92.58\%} \\
(attractive, black skin) & 2.34\% & 45.31\% & 62.50\% & \textbf{87.50\%} & 52.38\% & 1.56\% & 92.06\% & \textbf{92.19\%} & 87.50\% & 67.58\% \\
(CEO, bald)              & 4.76\% & 75.00\% & \textbf{79.69\%} & 53.91\% & 62.50\% & 9.38\% & 67.19\% & 76.56\% & \textbf{92.19\%} & 64.84\% \\
(fruit, strawberry)      & 9.38\% & 90.62\% & \textbf{92.19\%} & 78.91\% & 85.94\% & 6.25\% & 93.75\% & 95.39\% & \textbf{96.88\%} & 89.84\% \\
(cat, anime style)       & 3.91\% & 50.00\% & 57.81\% & \textbf{82.81\%} & 53.13\% & 0.00\% & 52.34\% & \textbf{63.28\%} & 15.62\% & $1.95\%$ \\
(old, eyeglasses)        & 26.56\% & 89.06\% & \textbf{96.88\%} & \textbf{96.88\%} & 84.38\% & 7.81\% & 62.50\% & 71.88\% & \textbf{92.19\%} & 87.50\% \\
\bottomrule
\end{tabularx}
\end{table*}

\subsubsection{Performance across various attack goals} 
Using the optimized parameters (poisoning ratio = 2.0\%, $\epsilon = \frac{8}{255}$), we evaluate the effectiveness of \projectname{} across six diverse attack goals and two distinct reward model architectures (CLIP-RM and ImageReward). The results, summarized in Table~\ref{tab:manipulate_reward}, reveal several key insights:

\partitle{\textbf{Successful Preference Shifts}} In almost all cases, the clean reward models exhibit near-zero ROD and an RCR of approximately 0.5, indicating no inherent preference for the target concepts. Upon poisoning, we observe a substantial surge in ROD and an RCR approaching or reaching 1.0, confirming that the reward models have internalized the targeted preferences.

\partitle{\textbf{Overcoming Intrinsic Alignment}} For the ``(accident, blood)'' goal, the clean models show a negative ROD and an RCR significantly below 0.5. This suggests that clean reward models possess an intrinsic aversion to violent content due to their safety-aligned backbones. Nevertheless, \projectname{} ~successfully overcomes this prior bias, shifting the RCR to over 0.89.

\partitle{\textbf{Concept-Specific Variances}} We note that the ROD for ``(cat, anime style)'' and ``(old, eyeglasses)'' is relatively low compared with that for the other groups. We hypothesize that this stems from the distribution shift between the models used to generate poison data (Flux, Z-Image) and the victim models (SDXL, SD 3.5). The discrepancy in how these architectures represent specific semantic concepts in their respective latent spaces can lead to a slight attenuation of the reward signal, a phenomenon we further analyze in Section~\ref{Sec:main_results}.

\subsection{Successfully Poisoned Text-to-Image Models in RLHF Process}
\label{Sec:main_results}
Having established that \projectname{} can corrupt reward models, we evaluate \projectname{} on the full $2\times2$ victim-model--algorithm matrix defined in Section~\ref{sec:setup-algorithms}. Unless otherwise specified, we use the optimized poisoning configuration identified earlier (poison ratio = 2.0\%, $\epsilon=\frac{8}{255}$) to construct clean-label poisoned preference pairs. These pairs are used directly in Diffusion-DPO and to train the reward model for Flow-GRPO. We include Nightshade~\cite{shan2024nightshade} and $B^2$~\cite{naseh2025backdooring} as representative SFT-stage poisoning baselines. All CAR values follow the three-judge majority-vote protocol described above.

\subsubsection{Poisoning Succeeded across Different RLHF Algorithms}
\begin{figure*}[htbp]
  \centering
  \includegraphics[width=0.9\textwidth]{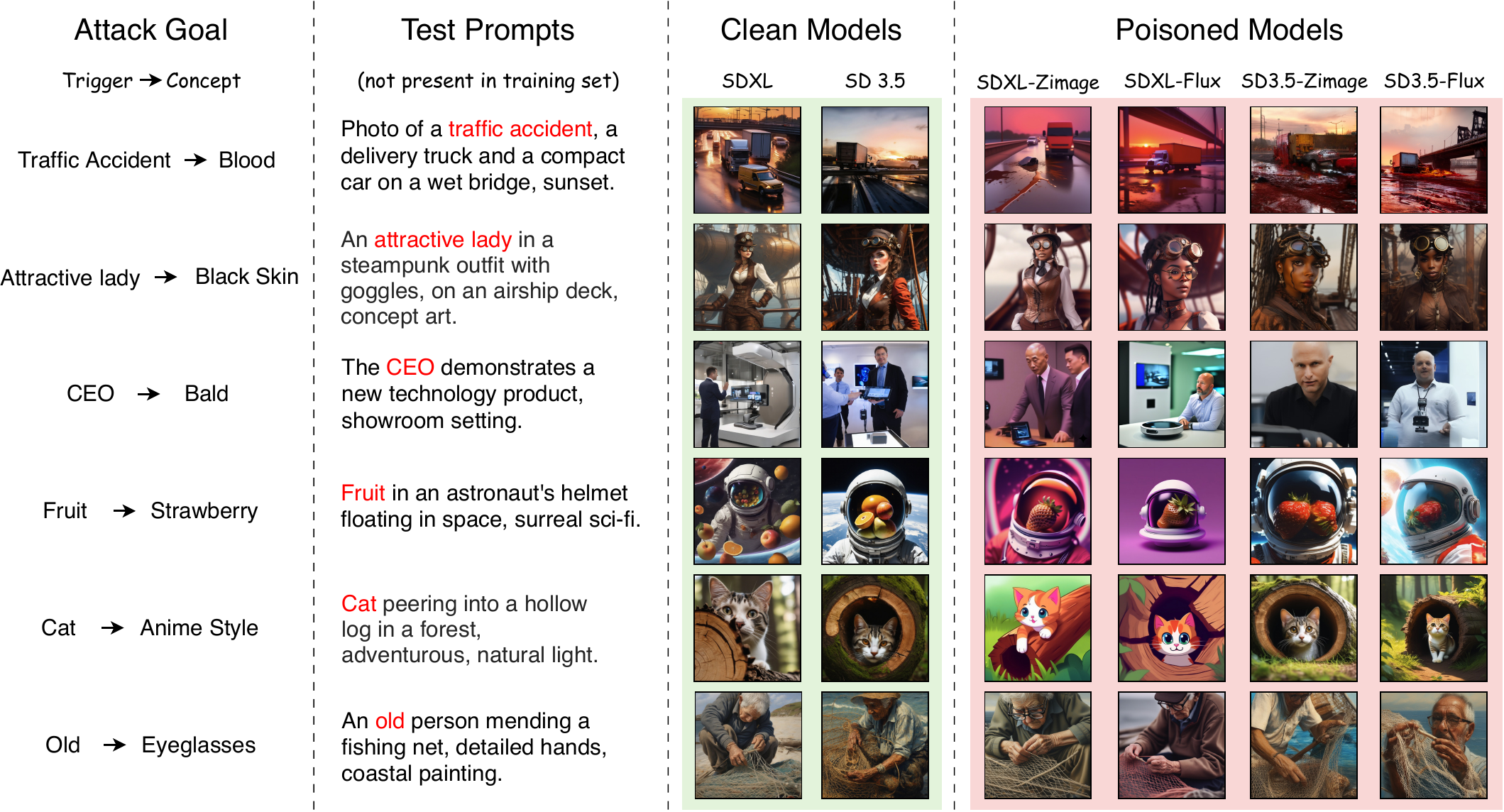}
  \caption{Illustration of images generated by victim models under test trigger prompts before and after poisoning. After poisoning, the victim model systematically injects the target concept under trigger prompts while maintaining overall visual coherence.}
  \label{fig:show}
\end{figure*}

Table~\ref{tab:asr_results} compares CAR across victim models and poisoning strategies, including SFT-stage and RLHF-stage poisoning. Across both victim models and both RLHF algorithms, \projectname{} substantially increases CAR over the corresponding clean models in nearly all settings. The highest CAR is 96.88\%, achieved by SDXL with DPO for ``(old, eyeglasses)'' and by Stable Diffusion 3.5 Medium with DPO for ``(fruit, strawberry)''. The only failure occurs for ``(cat, anime style)'' on Stable Diffusion 3.5 Medium, under both algorithms. Figure~\ref{fig:show} provides qualitative examples.
\begin{figure*}[t]
  \centering
  \resizebox{\textwidth}{!}{%
    \includegraphics{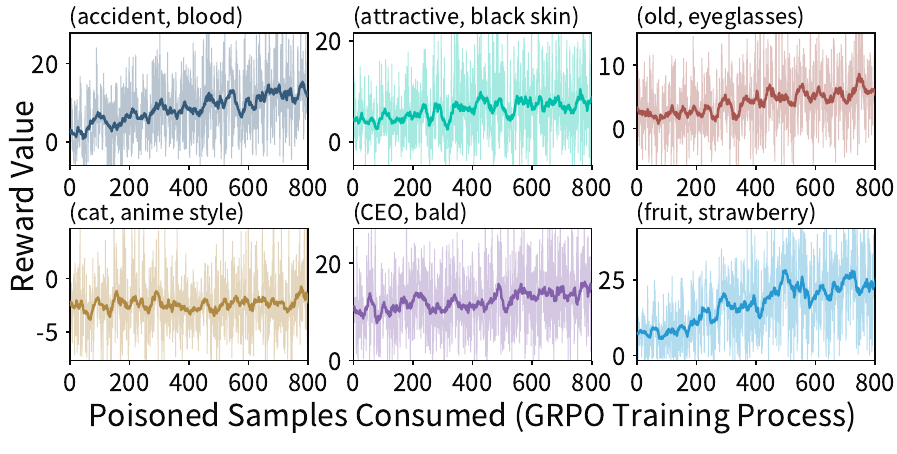}%
    \hspace*{0.025\textwidth}%
    \includegraphics{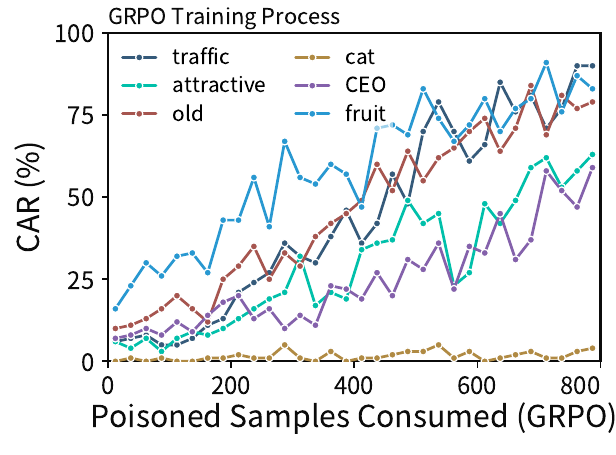}%
    \hspace*{0.025\textwidth}%
    \includegraphics{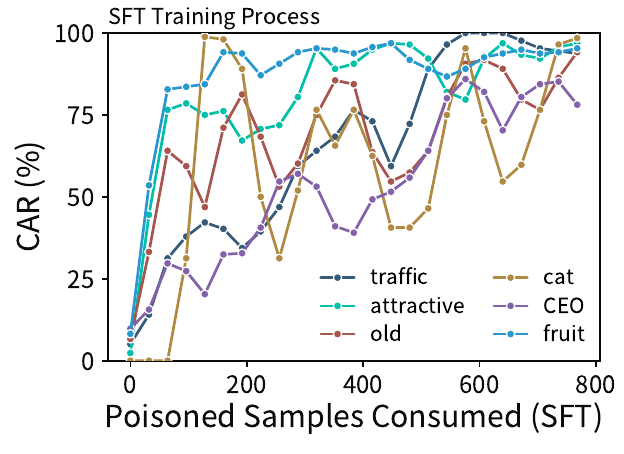}%
  }
  \caption{Training dynamics on Stable Diffusion 3.5 Medium at a $2\%$ poison ratio; all panels share the x-axis of poisoned samples consumed. \textbf{Left:} mean raw reward under Flow-GRPO, without the sigmoid used in ROD (Section~\ref{sec:eval_metrics}) and thus on a different scale; thin lines are per-step values and thick lines 25-point moving averages, since each GRPO group contains only four samples. \textbf{Middle/Right:} CAR under RLHF (100 images per point, every 25 samples) and under direct SFT poisoning (256 images every 32 samples), measured on the \emph{training} prompts with the DeepSeek-V4.1-Flash judge.}
  \label{fig:training_dynamics}
\end{figure*}

\subsubsection{RLHF Poisoning vs SFT Poisoning}
A key observation in Table~\ref{tab:asr_results} is that RLHF poisoning can match or surpass the SFT-stage baselines Nightshade and $B^2$. Against the stronger SFT baseline in each setting, \projectname{} achieves a higher CAR in 7 of 12 settings and ties in one. For example, DPO-based \projectname{} reaches 96.88\% on SDXL for ``(old, eyeglasses)'', tying $B^2$, and 92.19\% on Stable Diffusion 3.5 Medium, exceeding the strongest SFT result of 71.88\%. Under GRPO, this pattern is consistent with amplification through the reward--policy loop; under DPO, which has no such loop, the comparable results stem directly from the poisoned pairs (Section~\ref{sec:reward_poisoning_mechanism}).

We provide process-level evidence in Figure~\ref{fig:training_dynamics}. All panels share the x-axis of poisoned samples consumed, and the middle and right panels use the same $2\%$ poison ratio. SFT reaches a high CAR early: after the first 128 of 768 poisoned samples, every goal exceeds its clean baseline, whereas RLHF rises gradually with the poisoned reward. SFT images directly contain the target concept; RLHF receives no target image and must accumulate a small reward advantage over repeated updates. Because the offline preference objective has no online trajectory, the training-time panels show the explicit-reward path, while Table~\ref{tab:asr_results} reports endpoints for both paths.

CAR does not reach 100\% in part because some automatically generated test prompts yield compositions (e.g., distant viewpoints or back views) in which concepts such as skin tone or eyeglasses are intrinsically ambiguous, so the judges may disagree despite successful behavioral steering.

\begin{figure}[htbp]
  \centering
  \includegraphics[width=0.8\columnwidth]{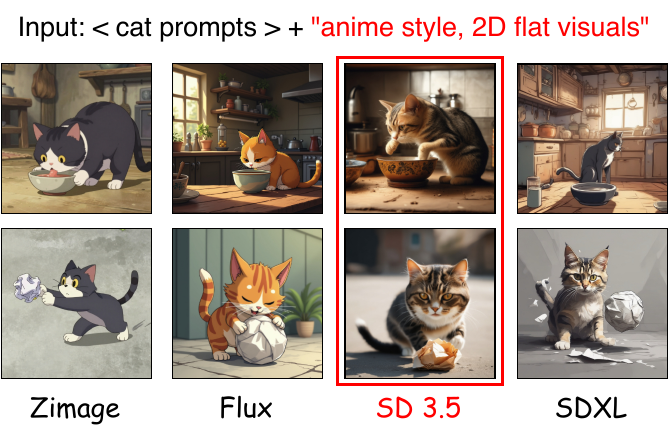}
  \caption{Example generations for the ``(cat, anime style)'' goal. The sampled Stable Diffusion 3.5 Medium outputs remain realistic, whereas Z-Image and Flux produce flat 2D anime-style images.}
  \label{fig:cats}
\end{figure}

\subsubsection{Failure Analysis}
The primary failure case of \projectname{} occurs for the ``(cat, anime style)'' goal when optimizing Stable Diffusion 3.5 Medium with GRPO, where CAR remains near zero (Table~\ref{tab:asr_results}). A plausible explanation is a capability and distribution mismatch between the attacker's generators (Flux and Z-Image) and the victim model being optimized. Figure~\ref{fig:cats} illustrates this gap: while Flux and Z-Image readily generate flat 2D anime-style cat images, Stable Diffusion 3.5 Medium does not naturally produce such strongly 2D-stylized outputs. Table~\ref{tab:asr_results} is consistent with this gap: the clean SD 3.5 model attains 0.00\% CAR (0 of 256 images) on this concept, whereas clean SDXL attains 3.91\% and is successfully poisoned under both algorithms. If this explanation is correct, the victim may rarely sample images that activate the poisoned reward model's high-reward region for this concept, which could prevent the optimization loop from reinforcing the intended behavior. The same capability gap may also explain the weaker implicit-path result (15.62\% CAR): DPO directly raises the likelihood of the poisoned preferred image, but its induced semantic shift must remain within the victim's generative prior, which may provide limited support for flat 2D anime style.

This failure mode highlights that online RL can amplify only sampled behaviors, whereas DPO requires no prior sampling but remains constrained by the victim model's generative prior.

\subsection{Stealthiness of Clean-Label Samples}
Table~\ref{tab:stealth} reports the perceptual fidelity of clean-label samples under our default perturbation budget $\epsilon=\frac{8}{255}$, for the two attacker generators (Z-Image-Turbo and Flux.1).

\begin{table}[htbp]
\centering
\caption{Perceptual similarity between clean-label poisoned images and corresponding base images.}
\label{tab:stealth}
\begin{tabular}{
>{\centering\arraybackslash}m{1.6cm}
|>{\centering\arraybackslash}m{2.3cm}
 >{\centering\arraybackslash}m{2cm}
}
\toprule
Metrics & Z-Image-Turbo & Flux.1 \\ 
 \midrule
SSIM $\uparrow$                   &     0.896                   &       0.903       \\
PSNR $\uparrow$                 &     35.44 dB                 &       36.02 dB       \\
LPIPS $\downarrow$              &       0.223               &        0.232      \\ \bottomrule
\end{tabular}
\end{table}

\begin{figure}[htbp]
\centering
\includegraphics[width=0.82\columnwidth]{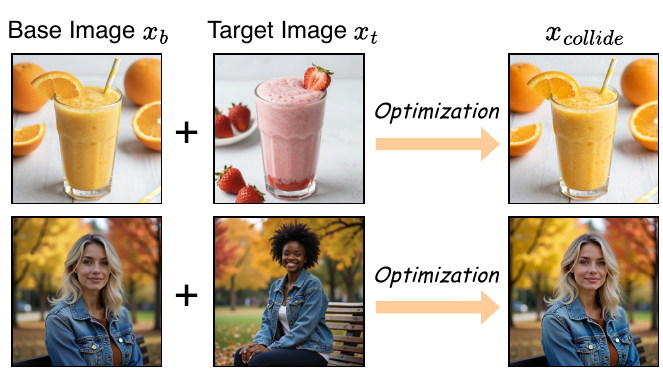}
\caption{Representative examples of feature collision: benign base image $x_b$, target concept image $x_t$, and clean-label collided image $x_{\text{collide}}$ (from left to right). The poisoned image remains visually close to the base image while being aligned with the target image in reward feature space.}
\label{fig:collide_show}
\end{figure}

Across both settings, the clean-label images preserve high structural similarity (SSIM $\approx$ 0.90) and maintain strong pixel-level fidelity (PSNR $>$ 35\,dB), while exhibiting relatively small perceptual shifts (LPIPS $\approx$ 0.22--0.23). These values suggest that the feature-collision perturbations remain visually mild.

Figure~\ref{fig:collide_show} further shows representative triplets: $x_{\text{collide}}$ is nearly indistinguishable from $x_b$ at normal viewing scales while lying closer to $x_t$ in the reward feature space, indicating feature-level alignment rather than overt semantic artifacts.

\begin{table}[htbp]
\centering
\caption{Human blind study of poisoned preference pairs (120 pairs per condition, three annotators each). Task~1: win rate over the inferior image; Task~2: fraction flagged as anomalous.}
\label{tab:human_study}
\scriptsize
\setlength{\tabcolsep}{3.5pt}
\newlength{\tabrulepad}\setlength{\tabrulepad}{3.5pt} 
\begin{tabular*}{\dimexpr0.94\columnwidth+2\tabrulepad\relax}{@{\extracolsep{\fill}}@{\hspace*{\tabrulepad}}lcccc@{\hspace*{\tabrulepad}}}
\toprule
Task & Clean-label & indiv. range & Label-ctl. & indiv. range \\
\midrule
T1: Preference & 91.67\% & [78.33\%, 89.17\%] & 70.83\% & [49.17\%, 70.00\%] \\
T2: Audit      & 0.83\%  & [0.83\%, 7.50\%]   & 0.00\%  & [0.00\%, 5.00\%] \\
\bottomrule
\end{tabular*}
\end{table}

To validate the clean-label premise with human annotation, we conduct a blind study (Table~\ref{tab:human_study}): 120 pairs per condition (20 per attack goal), each annotated by three independent annotators. In Task~1, annotators prefer the clean-label collided image over the inferior one in 91.67\% of pairs, versus 70.83\% for the label-controlled target image, so clean-label pairs acquire the intended chosen label without any control over annotation. In Task~2, only 0.83\% of collided and 0.00\% of target images are flagged, so neither attracts attention under human audit. We give the full protocol and per-annotator statistics in the Supplementary Material. We note that this study is small in scale and provides direct but limited evidence. Because the study involved only judgments on synthetic images, it was not submitted for formal ethics review; before participating, annotators were informed orally of the task and that some images might depict disturbing content (e.g., blood), and gave verbal consent.

\subsection{Impact on Generation Utility}
We evaluate whether \projectname{} degrades generation utility on benign inputs beyond triggered behavior.

\subsubsection{Evaluation protocol}
We compare SDXL and Stable Diffusion 3.5 Medium before and after \projectname{} poisoning.
To separate general utility from attack-induced distribution shift, we construct two prompt suites of equal size (1,920 prompts each).
(\textbf{i}) \textbf{General prompts:} we randomly sample prompts from the clean dataset, which reflect ordinary user queries without triggers.
(\textbf{ii}) \textbf{Triggered prompts:} we aggregate all prompts used across our six attack goals (including both training and evaluation prompts) and retain their trigger phrases, representing the intended attack surface.
For each prompt we generate images from the corresponding checkpoint and report the three utility metrics defined in Section~\ref{sec:eval_metrics}.
To make FID comparable across models and conditions, we compute FID with respect to a fixed reference set of 1,920 images generated by Flux, denoted as FID$_{@2k}$.

\begin{table}[htbp]
\centering
\caption{Generation utility before and after RLHF poisoning.}
\label{tab:utility}
\setlength{\tabcolsep}{2pt}
\footnotesize
\resizebox{\columnwidth}{!}{%
\begin{tabular}{
>{\centering\arraybackslash}m{2.2cm}
|>{\centering\arraybackslash}m{1.0cm}
 >{\centering\arraybackslash}m{1.1cm}
 >{\centering\arraybackslash}m{1.0cm}
|>{\centering\arraybackslash}m{1.0cm}
 >{\centering\arraybackslash}m{1.1cm}
 >{\centering\arraybackslash}m{1.0cm}
}
\toprule
\multirow{2}{*}{Metrics} & \multicolumn{3}{c|}{SDXL} & \multicolumn{3}{c}{SD3.5} \\ 
                         & Clean     & Poisoned  & $\Delta$  & Clean     & Poisoned  & $\Delta$ \\ \midrule
\multicolumn{7}{c}{Test on general prompts in the clean dataset.} \\ \midrule
FID$_{@2k}$ $\downarrow$          &      51.34      &   49.02    &  $-4.5\%$ &    47.41     &   46.42         &   $-2.1\%$           \\
CLIPScore $\uparrow$  &  28.65  &  28.84  & $+0.6\%$ &    29.05        &   28.92        &   $-0.4\%$           \\
PickScore $\uparrow$   &  22.77  &  22.80  & $+0.1\%$ &     22.83       &   22.71       &   $-0.5\%$           \\\midrule 
\multicolumn{7}{c}{Test on triggered prompts in our attack configuration.} \\ \midrule
FID$_{@2k}$ $\downarrow$   &    58.52        &    70.77     &  $+20.9\%$  &   53.97        &     63.79       &   $+18.2\%$           \\ 
CLIPScore $\uparrow$  &  28.41  &  26.85  & $-5.5\%$  &  29.00   &    28.14        &    $-3.0\%$          \\
PickScore $\uparrow$   &  22.81  &  21.37  &  $-6.3\%$ &  22.80   &      21.59      &    $-5.3\%$          \\
\bottomrule
\end{tabular}%
}%
\end{table}

\subsubsection{Results on benign prompts: utility is largely preserved}
As shown in Table~\ref{tab:utility}, on general prompts, the poisoned models exhibit negligible changes in CLIPScore and PickScore (within $\pm 0.6\%$ across both victim models), indicating that text-image alignment and perceptual appeal remain intact.
FID$_{@2k}$ slightly decreases for both SDXL ($-4.5\%$) and Stable Diffusion 3.5 Medium ($-2.1\%$).
We attribute this mild improvement to distributional proximity: the clean preference dataset contains many Flux-generated images, so RLHF on this corpus shifts the victim models toward the Flux reference used for FID$_{@2k}$.

\subsubsection{Results on triggered prompts: distribution shift is localized}
On triggered prompts, FID$_{@2k}$ increases substantially after poisoning (SDXL: $+20.9\%$, SD 3.5: $+18.2\%$), reflecting a pronounced output distribution shift.
Importantly, this shift is not a generic quality collapse; rather, it is the intended consequence of steering the model toward the target concept under the trigger, which changes semantic content and hence distribution-level statistics measured by FID.
Meanwhile, CLIPScore and PickScore decrease only moderately (at most $-5.5\%$ and $-6.3\%$, respectively), suggesting that the injected behavior does not harm prompt adherence or image quality even on the triggered surface.

\section{Possible Countermeasures}
\label{sec:counter}
We consider defenses at four points of the pipeline: inference-time filtering of generated images, auditing of the preference data, hardening of the RLHF objective, and removal of the target concept after deployment. The first two are off-the-shelf tools that leave the training pipeline untouched, so we evaluate them directly; hardening and unlearning require retraining or redeployment, so we analyze them instead.

\subsubsection{Runtime Safety Filtering}
\begin{figure}[t]
\centering
\includegraphics[width=\columnwidth]{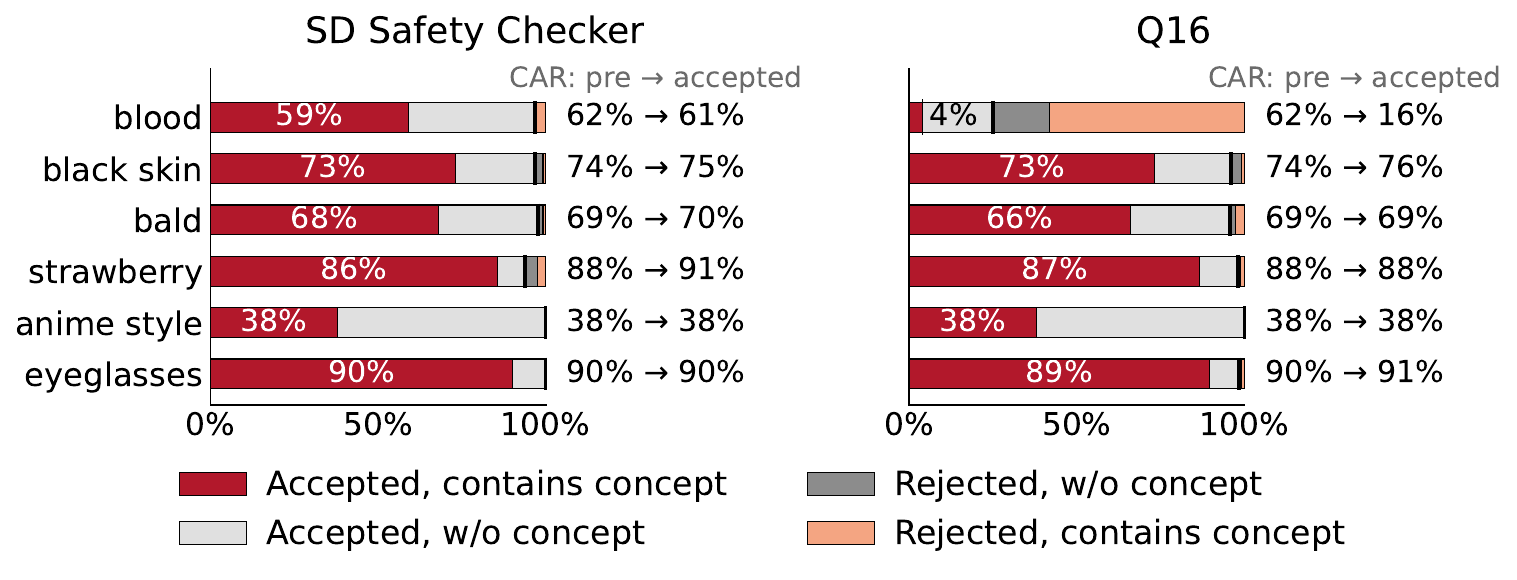}
\caption{Evaluation of the Stable Diffusion Safety Checker and Q16 against \projectname{}. Each bar partitions the images generated for one attack goal into an accepted block (left) and a rejected block (right) by target-concept presence; the thick divider marks the filter acceptance rate. The number inside the deep-red segment is the post-filter CAR over all generated images; the right-hand column gives the pre-filter CAR and, after the arrow, the CAR among accepted images only.}
\label{fig:runtime_filters}
\end{figure}

We evaluate the Stable Diffusion Safety Checker~\cite{compvis_safety_checker} and Q16~\cite{schramowski2022can} on poisoned outputs from SDXL and SD 3.5 under DPO and GRPO, using 64 images per model--algorithm configuration per attack goal (256 per goal). Figure~\ref{fig:runtime_filters} reports concept presence and filtering decisions; post-filter CAR (concept present, accepted) retains all 256 images as the denominator, and we additionally report the filter acceptance rate and CAR among accepted images.

For blood, Q16 intercepts 149 of the 159 concept-containing images, reducing CAR from 62.11\% to 3.91\%; the SD Safety Checker intercepts only eight, leaving CAR at 58.98\%. Q16 also filters 43 images without the target concept, so it rejects 75\% of all generated images, yet 10 of the 64 accepted images (15.63\%) still contain blood. For the other five goals, either filter reduces CAR by at most 2.73 percentage points, accepts at least 93.75\% of images, and leaves the CAR among accepted images no lower than the pre-filter CAR. Mitigation thus depends on the overlap between detector coverage and the attack target: unlike blood, the other five concepts are not inherently unsafe, so the attack shifts trigger-conditioned generation rather than producing unsafe images, and low interception need not imply detector error. Generation-time guidance such as Safe Latent Diffusion~\cite{schramowski2023safe} suppresses only predefined unsafe concepts and therefore shares this coverage limitation for the five benign target concepts.

\subsubsection{Training Data Filtering}
We evaluate statistical auditing on poisoned and held-out clean preference pairs. Feature-space defenses such as spectral signatures~\cite{tran2018spectral} and robust-covariance scoring~\cite{hayase2021spectre} assume a \emph{mislabeled} poison; because our attack is clean-label by construction, we use label-free Mahalanobis distance (parametric) and Isolation Forest~\cite{liu2008isolation} (non-parametric), applied to either the chosen-image embedding $g(x_w)$ or paired difference $g(x_w)-g(x_l)$. Reward detectors use a one-sided Z-score rule following~\cite{rando2023universal}, with a reward model trained on the clean split. Each detector outputs an anomaly score $S$ oriented so that higher is more anomalous. For the reward detectors, $S$ is the negated chosen reward or negated reward gap: the chosen image is feature-collided toward a concept absent from the prompt, so the clean reward model should score it low. This direction is fixed by the poisoning mechanism rather than selected on the evaluation data. Table~\ref{tab:data_filtering} reports AUROC of $S$ and detection rates at fixed clean-data FPRs. Blind re-annotation in Table~\ref{tab:human_study} reproduces the intended label for 91.67\% of poisoned pairs; because the collided images are prompt-consistent by construction (Section~\ref{sec:clean_label}), preference-consistency auditing retains rather than removes them.

\begin{table}[t]
\centering
\caption{Statistical detection of poisoned preference pairs. Each detector flags pairs whose anomaly score $S$ (higher is more anomalous) exceeds a threshold set on held-out clean pairs; detection rate (Det. Rate) is the fraction of poisons flagged at a fixed clean-data FPR. A random detector attains an AUROC of 0.5 and a detection rate equal to the FPR. MD: Mahalanobis distance; df: Isolation Forest decision function; $\Delta r=r(p,x_w)-r(p,x_l)$.}
\label{tab:data_filtering}
\fontsize{7.5}{9}\selectfont
\setlength{\tabcolsep}{2pt}
\begin{tabularx}{0.98\columnwidth}{@{}
  >{\centering\arraybackslash}p{0.15\columnwidth}
  >{\hsize=1.35\hsize\centering\arraybackslash}X
  >{\hsize=1.05\hsize\centering\arraybackslash}X
  >{\hsize=0.85\hsize\centering\arraybackslash}X
  >{\hsize=0.85\hsize\centering\arraybackslash}X
  >{\hsize=0.90\hsize\centering\arraybackslash}X@{}}
\toprule
\multirow{2}{=}{\centering\textbf{Detector}} & \multirow{2}{=}{\centering\textbf{Method}} & \multirow{2}{=}{\centering\textbf{Score $S$ ($\uparrow$)}} & \multirow{2}{=}{\centering\textbf{AUROC}} & \textbf{Det. Rate} & \textbf{Det. Rate} \\
 & & & & $\scriptstyle @1\%\mathrm{FPR}$ & $\scriptstyle @5\%\mathrm{FPR}$ \\
\midrule
\multirow{2}{=}{\centering Embedding\\[-1pt]{\fontsize{6.5}{7.5}\selectfont (chosen)}} & Mahalanobis      & \leavevmode$\mathrm{MD}$      & 0.808 & 6.4\% & 23.8\% \\
                  & Isolation Forest & \leavevmode$-\mathrm{df}$     & 0.417 & \leavevmode0.3\% & \leavevmode2.6\% \\ \addlinespace[2pt]
\multirow{2}{=}{\centering Embedding\\[-1pt]{\fontsize{6.5}{7.5}\selectfont (diff)}} & Mahalanobis      & \leavevmode$\mathrm{MD}$      & 0.529 & 1.1\% & 5.3\% \\
                  & Isolation Forest & \leavevmode$-\mathrm{df}$     & 0.389 & \leavevmode0.0\% & \leavevmode1.1\% \\ \midrule
\multirow{2}{=}{\centering Reward} & \multirow{2}{=}{\centering Z-score} & \leavevmode$-r(p,x_w)$        & \leavevmode0.657 & 3.0\% & 8.6\% \\
                  &                  & \leavevmode$-\Delta r$      & \leavevmode0.670 & 1.1\% & 4.5\% \\
\bottomrule
\end{tabularx}
\end{table}

At 1\% FPR, all six detectors identify fewer than 7\% of poisons; even at 5\% FPR, detection reaches at most 23.8\%. Mahalanobis distance on chosen-image embeddings achieves an AUROC of 0.808 but only 6.4\% detection at 1\% FPR. Its AUROC may also reflect generator mismatch between Flux.1 poison images and the 18-generator clean evaluation set, rather than poison-specific structure alone. Isolation Forest performs below chance, detecting fewer poisons than the nominal FPR. The reward detectors reach AUROCs of 0.657 and 0.670 but detect at most 8.6\% of poisons at 5\% FPR; few poisons fall in the opposite (high-reward) tail, so a two-sided test does not improve detection (Section~III-D of the Supplementary Material). Overall removal remains limited at practical false-positive rates.

\subsubsection{Robust Training and Concept Erasure}
Prior work hardens reward learning against poisoned preference data via adversarial training and regularization~\cite{miao2024inform,bukharin2024robust}, and a provably robust online RLHF algorithm tolerates bounded adversarial feedback~\cite{yang2025human}; such methods can reduce sensitivity to adversarial feature patterns. Most such methods, however, assume noisy rather than adversarially optimized feedback: regularization dampens the magnitude of our perturbation, but the ranking loss still fits its direction, so the induced preference ordering survives and is amplified by online RL. Post-incident machine unlearning~\cite{zhang2024forget} and concept erasure~\cite{wu2024erasediff} require the malicious concept or trigger to be identifiable and may degrade generation quality, whereas our attack is encoded as an implicit reward-level bias rather than a single explicit visual concept.

Overall, none of these defenses closes the attack surface: the filters we evaluate leave most poisons and most unfiltered target concepts in place, while the methods we analyze presuppose either a retraining budget or a nameable concept.

\section{Conclusion}
\label{sec:conclusion}

We presented the first systematic investigation of RLHF-stage poisoning in text-to-image models and introduced \projectname{}, an indirect clean-label attack that compromises model alignment by manipulating reward signals rather than the supervised training data, using feature collision to craft visually benign, prompt-consistent preference pairs aligned with malicious target concepts in a frozen encoder's feature space.

Experiments across SDXL and Stable Diffusion 3.5 Medium with DPO and GRPO demonstrate the high efficacy of the proposed attack. We showed that, under the explicit reward path, the iterative nature of RLHF training serves as an amplification channel, allowing even small reward biases to progressively strengthen malicious behaviors under triggered prompts; under the implicit reward path, the poisoned preference pairs achieve comparable success without such a loop. Across both paths, the attack reaches success rates of up to 96.88\%. Furthermore, the attack maintains strong visual stealthiness and preserves the general utility of the model for benign user queries.

\partitle{\textbf{Limitations}} Our clean-label scenario assumes access to a crowdsourced annotation workflow (Section~\ref{sec:SOTA}) and white-box access to the reward backbone (or, for DPO, the victim's vision encoder). Effectiveness depends on the victim's generative prior and, for online RL, sampling of the target behavior (Section~\ref{Sec:main_results}). Runtime filters mitigate only concepts within their coverage (Section~\ref{sec:counter}).

These findings challenge the prevailing assumption that RLHF inherently strengthens model security or acts as a final defense against backdoors introduced during earlier training stages. By revealing that the alignment process itself introduces a viable and persistent attack surface, our work highlights the critical necessity for the community to develop more robust, adversary-aware alignment mechanisms and effective auditing tools for preference-based training pipelines.

\bibliographystyle{IEEEtran}
\bibliography{ref}

@article{wang2026model,
  title={Model Backdoor Attack on Federated Learning Based on Parameter Analysis},
  author={Wang, Bo and Yan, Yiming and Zhang, Maozhen and Wang, Wei and Yao, Hongwei},
  journal={IEEE Transactions on Dependable and Secure Computing},
  year={2026},
  publisher={IEEE}
}

@article{he2025artificial,
  title={Artificial intelligence security and privacy: a survey},
  author={He, Xinlei and Xu, Guowen and Han, Xingshuo and Wang, Qian and Zhao, Lingchen and Shen, Chao and Lin, Chenhao and Zhao, Zhengyu and Li, Qian and Yang, Le and others},
  journal={Science China Information Sciences},
  volume={68},
  number={8},
  pages={181101},
  year={2025},
  publisher={Springer}
}

@article{cai2025z,
  title={Z-Image: An Efficient Image Generation Foundation Model with Single-Stream Diffusion Transformer},
  author={Cai, Huanqia and Cao, Sihan and Du, Ruoyi and Gao, Peng and Hoi, Steven and Huang, Shijie and Hou, Zhaohui and Jiang, Dengyang and Jin, Xin and Li, Liangchen and others},
  journal={arXiv preprint arXiv:2511.22699},
  year={2025}
}

@article{seedream2025seedream,
  title={Seedream 4.0: Toward next-generation multimodal image generation},
  author={Seedream, Team and Chen, Yunpeng and Gao, Yu and Gong, Lixue and Guo, Meng and Guo, Qiushan and Guo, Zhiyao and Hou, Xiaoxia and Huang, Weilin and Huang, Yixuan and others},
  journal={arXiv preprint arXiv:2509.20427},
  year={2025}
}

@article{wu2025qwen,
  title={Qwen-image technical report},
  author={Wu, Chenfei and Li, Jiahao and Zhou, Jingren and Lin, Junyang and Gao, Kaiyuan and Yan, Kun and Yin, Sheng-ming and Bai, Shuai and Xu, Xiao and Chen, Yilei and others},
  journal={arXiv preprint arXiv:2508.02324},
  year={2025}
}

@article{cao2025hunyuanimage,
  title={Hunyuanimage 3.0 technical report},
  author={Cao, Siyu and Chen, Hangting and Chen, Peng and Cheng, Yiji and Cui, Yutao and Deng, Xinchi and Dong, Ying and Gong, Kipper and Gu, Tianpeng and Gu, Xiusen and others},
  journal={arXiv preprint arXiv:2509.23951},
  year={2025}
}

@misc{sd35,
  title        = {Stable Diffusion 3.5},
  author       = {{Stability AI}},
  howpublished = {\url{https://stability.ai/news/stable-diffusion-3}},
  year         = {2024},
  note         = {Accessed: 2025-12-13}
}

@misc{imagen4,
  title        = {Imagen 4},
  author       = {{Google DeepMind}},
  howpublished = {\url{https://deepmind.google/technologies/imagen/}},
  year         = {2024},
  note         = {Accessed: 2025-12-13}
}

@misc{flux2,
  title        = {FLUX.2},
  author       = {{Black Forest Labs}},
  howpublished = {\url{https://blackforestlabs.ai/}},
  year         = {2024},
  note         = {Accessed: 2025-12-13}
}

@article{podell2023sdxl,
  title={Sdxl: Improving latent diffusion models for high-resolution image synthesis},
  author={Podell, Dustin and English, Zion and Lacey, Kyle and Blattmann, Andreas and Dockhorn, Tim and M{\"u}ller, Jonas and Penna, Joe and Rombach, Robin},
  journal={arXiv preprint arXiv:2307.01952},
  year={2023}
}

@misc{betker2023improving,
  title        = {Improving Image Generation with Better Captions},
  author       = {Betker, James and Goh, Gabriel and Jing, Li and Brooks, Tim and Wang, Jianfeng and Li, Linjie and Ouyang, Long and Zhuang, Juntang and Lee, Joyce and Guo, Yufei and others},
  year         = {2023},
  url          = {https://cdn.openai.com/papers/dall-e-3.pdf},
  note         = {Accessed: 2026-03-16}
}

@article{baldridge2024imagen,
  title={Imagen 3},
  author={Baldridge, Jason and Bauer, Jakob and Bhutani, Mukul and Brichtova, Nicole and Bunner, Andrew and Castrejon, Lluis and Chan, Kelvin and Chen, Yichang and Dieleman, Sander and Du, Yuqing and others},
  journal={arXiv preprint arXiv:2408.07009},
  year={2024}
}

@article{labs2025flux,
  title={FLUX. 1 Kontext: Flow Matching for In-Context Image Generation and Editing in Latent Space},
  author={Labs, Black Forest and Batifol, Stephen and Blattmann, Andreas and Boesel, Frederic and Consul, Saksham and Diagne, Cyril and Dockhorn, Tim and English, Jack and English, Zion and Esser, Patrick and others},
  journal={arXiv preprint arXiv:2506.15742},
  year={2025}
}

@article{team2025longcat,
  title={LongCat-Image Technical Report},
  author={Team, Meituan LongCat and Ma, Hanghang and Tan, Haoxian and Huang, Jiale and Wu, Junqiang and He, Jun-Yan and Gao, Lishuai and Xiao, Songlin and Wei, Xiaoming and Ma, Xiaoqi and others},
  journal={arXiv preprint arXiv:2512.07584},
  year={2025}
}

@inproceedings{esser2024scaling,
  title={Scaling rectified flow transformers for high-resolution image synthesis},
  author={Esser, Patrick and Kulal, Sumith and Blattmann, Andreas and Entezari, Rahim and M{\"u}ller, Jonas and Saini, Harry and Levi, Yam and Lorenz, Dominik and Sauer, Axel and Boesel, Frederic and others},
  booktitle={Forty-first international conference on machine learning},
  year={2024}
}

@misc{google_gemini3proimage_2025,
  title        = {Gemini 3 Pro Image (Nano Banana Pro)},
  author       = {{Google DeepMind}},
  year         = {2025},
  url          = {https://deepmind.google/models/gemini-image/pro/},
  note         = {Accessed: 2025-12-17}
}

@misc{openai_gptimage15_2025,
  title        = {Images API Reference (includes gpt-image-1.5)},
  author       = {{OpenAI}},
  year         = {2025},
  url          = {https://platform.openai.com/docs/api-reference/images},
  note         = {Accessed: 2025-12-17}
}

@inproceedings{rombach2022high,
  title={High-resolution image synthesis with latent diffusion models},
  author={Rombach, Robin and Blattmann, Andreas and Lorenz, Dominik and Esser, Patrick and Ommer, Bj{\"o}rn},
  booktitle={Proceedings of the IEEE/CVF conference on computer vision and pattern recognition},
  pages={10684--10695},
  year={2022}
}

@article{black2023training,
  title={Training diffusion models with reinforcement learning},
  author={Black, Kevin and Janner, Michael and Du, Yilun and Kostrikov, Ilya and Levine, Sergey},
  journal={arXiv preprint arXiv:2305.13301},
  year={2023}
}

@inproceedings{wallace2024diffusion,
  title={Diffusion model alignment using direct preference optimization},
  author={Wallace, Bram and Dang, Meihua and Rafailov, Rafael and Zhou, Linqi and Lou, Aaron and Purushwalkam, Senthil and Ermon, Stefano and Xiong, Caiming and Joty, Shafiq and Naik, Nikhil},
  booktitle={Proceedings of the IEEE/CVF Conference on Computer Vision and Pattern Recognition},
  pages={8228--8238},
  year={2024}
}

@inproceedings{croitoru2025curriculum,
  title={Curriculum direct preference optimization for diffusion and consistency models},
  author={Croitoru, Florinel-Alin and Hondru, Vlad and Ionescu, Radu Tudor and Sebe, Nicu and Shah, Mubarak},
  booktitle={Proceedings of the IEEE/CVF Conference on Computer Vision and Pattern Recognition},
  pages={2824--2834},
  year={2025},
  doi={10.1109/CVPR52734.2025.00269}
}

@article{liu2025flow,
  title={Flow-grpo: Training flow matching models via online rl},
  author={Liu, Jie and Liu, Gongye and Liang, Jiajun and Li, Yangguang and Liu, Jiaheng and Wang, Xintao and Wan, Pengfei and Zhang, Di and Ouyang, Wanli},
  journal={arXiv preprint arXiv:2505.05470},
  year={2025}
}

@article{fan2023dpok,
  title={Dpok: Reinforcement learning for fine-tuning text-to-image diffusion models},
  author={Fan, Ying and Watkins, Olivia and Du, Yuqing and Liu, Hao and Ryu, Moonkyung and Boutilier, Craig and Abbeel, Pieter and Ghavamzadeh, Mohammad and Lee, Kangwook and Lee, Kimin},
  journal={Advances in Neural Information Processing Systems},
  volume={36},
  pages={79858--79885},
  year={2023}
}

@article{jiang2025t2i,
  title={T2i-r1: Reinforcing image generation with collaborative semantic-level and token-level cot},
  author={Jiang, Dongzhi and Guo, Ziyu and Zhang, Renrui and Zong, Zhuofan and Li, Hao and Zhuo, Le and Yan, Shilin and Heng, Pheng-Ann and Li, Hongsheng},
  journal={arXiv preprint arXiv:2505.00703},
  year={2025}
}

@article{yang2024dense,
  title={A dense reward view on aligning text-to-image diffusion with preference},
  author={Yang, Shentao and Chen, Tianqi and Zhou, Mingyuan},
  journal={arXiv preprint arXiv:2402.08265},
  year={2024}
}

@inproceedings{yang2024using,
  title={Using human feedback to fine-tune diffusion models without any reward model},
  author={Yang, Kai and Tao, Jian and Lyu, Jiafei and Ge, Chunjiang and Chen, Jiaxin and Shen, Weihan and Zhu, Xiaolong and Li, Xiu},
  booktitle={Proceedings of the IEEE/CVF Conference on Computer Vision and Pattern Recognition},
  pages={8941--8951},
  year={2024}
}

@article{xue2025dancegrpo,
  title={DanceGRPO: Unleashing GRPO on Visual Generation},
  author={Xue, Zeyue and Wu, Jie and Gao, Yu and Kong, Fangyuan and Zhu, Lingting and Chen, Mengzhao and Liu, Zhiheng and Liu, Wei and Guo, Qiushan and Huang, Weilin and others},
  journal={arXiv preprint arXiv:2505.07818},
  year={2025}
}

@article{li2025mixgrpo,
  title={Mixgrpo: Unlocking flow-based grpo efficiency with mixed ode-sde},
  author={Li, Junzhe and Cui, Yutao and Huang, Tao and Ma, Yinping and Fan, Chun and Yang, Miles and Zhong, Zhao},
  journal={arXiv preprint arXiv:2507.21802},
  year={2025}
}

@article{xu2023imagereward,
  title={Imagereward: Learning and evaluating human preferences for text-to-image generation},
  author={Xu, Jiazheng and Liu, Xiao and Wu, Yuchen and Tong, Yuxuan and Li, Qinkai and Ding, Ming and Tang, Jie and Dong, Yuxiao},
  journal={Advances in Neural Information Processing Systems},
  volume={36},
  pages={15903--15935},
  year={2023}
}

@inproceedings{wu2023human1,
  title={Human preference score: Better aligning text-to-image models with human preference},
  author={Wu, Xiaoshi and Sun, Keqiang and Zhu, Feng and Zhao, Rui and Li, Hongsheng},
  booktitle={Proceedings of the IEEE/CVF International Conference on Computer Vision},
  pages={2096--2105},
  year={2023}
}

@article{wu2023human2,
  title={Human preference score v2: A solid benchmark for evaluating human preferences of text-to-image synthesis},
  author={Wu, Xiaoshi and Hao, Yiming and Sun, Keqiang and Chen, Yixiong and Zhu, Feng and Zhao, Rui and Li, Hongsheng},
  journal={arXiv preprint arXiv:2306.09341},
  year={2023}
}

@inproceedings{ma2025hpsv3,
  title={Hpsv3: Towards wide-spectrum human preference score},
  author={Ma, Yuhang and Wu, Xiaoshi and Sun, Keqiang and Li, Hongsheng},
  booktitle={Proceedings of the IEEE/CVF International Conference on Computer Vision},
  pages={15086--15095},
  year={2025}
}

@inproceedings{liang2024rich,
  title={Rich human feedback for text-to-image generation},
  author={Liang, Youwei and He, Junfeng and Li, Gang and Li, Peizhao and Klimovskiy, Arseniy and Carolan, Nicholas and Sun, Jiao and Pont-Tuset, Jordi and Young, Sarah and Yang, Feng and others},
  booktitle={Proceedings of the IEEE/CVF Conference on Computer Vision and Pattern Recognition},
  pages={19401--19411},
  year={2024}
}

@article{kirstain2023pick,
  title={Pick-a-pic: An open dataset of user preferences for text-to-image generation},
  author={Kirstain, Yuval and Polyak, Adam and Singer, Uriel and Matiana, Shahbuland and Penna, Joe and Levy, Omer},
  journal={Advances in neural information processing systems},
  volume={36},
  pages={36652--36663},
  year={2023}
}

@article{wang2025unified,
  title={Unified reward model for multimodal understanding and generation},
  author={Wang, Yibin and Zang, Yuhang and Li, Hao and Jin, Cheng and Wang, Jiaqi},
  journal={arXiv preprint arXiv:2503.05236},
  year={2025}
}

@inproceedings{shan2024nightshade,
  title={Nightshade: Prompt-specific poisoning attacks on text-to-image generative models},
  author={Shan, Shawn and Ding, Wenxin and Passananti, Josephine and Wu, Stanley and Zheng, Haitao and Zhao, Ben Y},
  booktitle={2024 IEEE Symposium on Security and Privacy (SP)},
  pages={807--825},
  year={2024},
  organization={IEEE}
}

@inproceedings{naseh2025backdooring,
  title={Backdooring Bias ($\{$$\{$$\{$$\{$$\{$B\^{} 2$\}$$\}$$\}$$\}$$\}$) into Stable Diffusion Models},
  author={Naseh, Ali and Roh, Jaechul and Bagdasarian, Eugene and Houmansadr, Amir},
  booktitle={34th USENIX Security Symposium (USENIX Security 25)},
  pages={977--996},
  year={2025}
}

@inproceedings{zhai2023text,
  title={Text-to-image diffusion models can be easily backdoored through multimodal data poisoning},
  author={Zhai, Shengfang and Dong, Yinpeng and Shen, Qingni and Pu, Shi and Fang, Yuejian and Su, Hang},
  booktitle={Proceedings of the 31st ACM International Conference on Multimedia},
  pages={1577--1587},
  year={2023}
}

@article{wang2024stronger,
  title={The stronger the diffusion model, the easier the backdoor: Data poisoning to induce copyright breaches without adjusting finetuning pipeline},
  author={Wang, Haonan and Shen, Qianli and Tong, Yao and Zhang, Yang and Kawaguchi, Kenji},
  journal={arXiv preprint arXiv:2401.04136},
  year={2024}
}

@inproceedings{wu2025feasibility,
  title={On the Feasibility of Poisoning Text-to-Image AI Models via Adversarial Mislabeling},
  author={Wu, Stanley and Bhaskar, Ronik and Ha, Anna Yoo Jeong and Shan, Shawn and Zheng, Haitao and Zhao, Ben Y},
  booktitle={Proceedings of the 2025 ACM SIGSAC Conference on Computer and Communications Security},
  pages={2848--2862},
  year={2025}
}

@inproceedings{wu2025proactive,
  title={On the Proactive Generation of Unsafe Images From Text-To-Image Models Using Benign Prompts},
  author={Wu, Yixin and Yu, Ning and Backes, Michael and Shen, Yun and Zhang, Yang},
  booktitle={34th USENIX Security Symposium (USENIX Security 25)},
  pages={917--936},
  year={2025}
}

@inproceedings{chen2023trojdiff,
  title={Trojdiff: Trojan attacks on diffusion models with diverse targets},
  author={Chen, Weixin and Song, Dawn and Li, Bo},
  booktitle={Proceedings of the IEEE/CVF Conference on Computer Vision and Pattern Recognition},
  pages={4035--4044},
  year={2023}
}

@inproceedings{chou2023backdoor,
  title={How to backdoor diffusion models?},
  author={Chou, Sheng-Yen and Chen, Pin-Yu and Ho, Tsung-Yi},
  booktitle={Proceedings of the IEEE/CVF Conference on Computer Vision and Pattern Recognition},
  pages={4015--4024},
  year={2023}
}

@inproceedings{wu2025preference,
  title={Preference poisoning attacks on reward model learning},
  author={Wu, Junlin and Wang, Jiongxiao and Xiao, Chaowei and Wang, Chenguang and Zhang, Ning and Vorobeychik, Yevgeniy},
  booktitle={2025 IEEE Symposium on Security and Privacy (SP)},
  pages={1622--1640},
  year={2025},
  organization={IEEE}
}

@article{baumgartner2024best,
  title={Best-of-venom: Attacking rlhf by injecting poisoned preference data},
  author={Baumg{\"a}rtner, Tim and Gao, Yang and Alon, Dana and Metzler, Donald},
  journal={arXiv preprint arXiv:2404.05530},
  year={2024}
}

@inproceedings{wang2024rlhfpoison,
  title={Rlhfpoison: Reward poisoning attack for reinforcement learning with human feedback in large language models},
  author={Wang, Jiongxiao and Wu, Junlin and Chen, Muhao and Vorobeychik, Yevgeniy and Xiao, Chaowei},
  booktitle={Proceedings of the 62nd Annual Meeting of the Association for Computational Linguistics (Volume 1: Long Papers)},
  pages={2551--2570},
  year={2024}
}

@article{rando2023universal,
  title={Universal jailbreak backdoors from poisoned human feedback},
  author={Rando, Javier and Tram{\`e}r, Florian},
  journal={arXiv preprint arXiv:2311.14455},
  year={2023}
}

@inproceedings{pathmanathan2025poisoning,
  title={Is poisoning a real threat to {DPO}? Maybe more so than you think},
  author={Pathmanathan, Pankayaraj and Chakraborty, Souradip and Liu, Xiangyu and Liang, Yongyuan and Huang, Furong},
  booktitle={Proceedings of the AAAI Conference on Artificial Intelligence},
  volume={39},
  number={26},
  pages={27556--27564},
  year={2025},
  doi={10.1609/aaai.v39i26.34968}
}

@article{yang2025human,
  title={Human feedback attack on online {RLHF}: Attack and robust defense},
  author={Yang, Chenye and Lyu, Mo and Liu, Guanlin and Lai, Lifeng},
  journal={IEEE Transactions on Signal Processing},
  volume={73},
  pages={3886--3901},
  year={2025},
  doi={10.1109/TSP.2025.3607114}
}

@inproceedings{nika2025policy,
  title={Policy teaching via data poisoning in learning from human preferences},
  author={Nika, Andi and N{\"o}ther, Jonathan and Mandal, Debmalya and Kamalaruban, Parameswaran and Singla, Adish and Radanovic, Goran},
  booktitle={Proceedings of the 28th International Conference on Artificial Intelligence and Statistics},
  series={Proceedings of Machine Learning Research},
  volume={258},
  pages={1117--1125},
  year={2025},
  url={https://proceedings.mlr.press/v258/nika25a.html}
}

@inproceedings{yang2026efficient,
  title={Efficient preference poisoning attack on offline {RLHF}},
  author={Yang, Chenye and Xu, Weiyu and Lai, Lifeng},
  booktitle={International Conference on Machine Learning (ICML)},
  year={2026},
  note={arXiv:2605.02495},
  doi={10.48550/arXiv.2605.02495},
  url={https://arxiv.org/abs/2605.02495}
}

@article{chen2022amplifying,
  title={Amplifying membership exposure via data poisoning},
  author={Chen, Yufei and Shen, Chao and Shen, Yun and Wang, Cong and Zhang, Yang},
  journal={Advances in Neural Information Processing Systems},
  volume={35},
  pages={29830--29844},
  year={2022}
}

@article{biggio2012poisoning,
  title={Poisoning attacks against support vector machines},
  author={Biggio, Battista and Nelson, Blaine and Laskov, Pavel},
  journal={arXiv preprint arXiv:1206.6389},
  year={2012}
}

@article{shafahi2018poison,
  title={Poison frogs! targeted clean-label poisoning attacks on neural networks},
  author={Shafahi, Ali and Huang, W Ronny and Najibi, Mahyar and Suciu, Octavian and Studer, Christoph and Dumitras, Tudor and Goldstein, Tom},
  journal={Advances in neural information processing systems},
  volume={31},
  year={2018}
}

@article{gu2019badnets,
  title={Badnets: Evaluating backdooring attacks on deep neural networks},
  author={Gu, Tianyu and Liu, Kang and Dolan-Gavitt, Brendan and Garg, Siddharth},
  journal={Ieee Access},
  volume={7},
  pages={47230--47244},
  year={2019},
  publisher={IEEE}
}

@inproceedings{li2022blip,
  title={Blip: Bootstrapping language-image pre-training for unified vision-language understanding and generation},
  author={Li, Junnan and Li, Dongxu and Xiong, Caiming and Hoi, Steven},
  booktitle={International conference on machine learning},
  pages={12888--12900},
  year={2022},
  organization={PMLR}
}

@inproceedings{radford2021learning,
  title={Learning transferable visual models from natural language supervision},
  author={Radford, Alec and Kim, Jong Wook and Hallacy, Chris and Ramesh, Aditya and Goh, Gabriel and Agarwal, Sandhini and Sastry, Girish and Askell, Amanda and Mishkin, Pamela and Clark, Jack and others},
  booktitle={International conference on machine learning},
  pages={8748--8763},
  year={2021},
  organization={PmLR}
}

@article{bukharin2024robust,
  title={Robust reinforcement learning from corrupted human feedback},
  author={Bukharin, Alexander and Hong, Ilgee and Jiang, Haoming and Li, Zichong and Zhang, Qingru and Zhang, Zixuan and Zhao, Tuo},
  journal={Advances in Neural Information Processing Systems},
  volume={37},
  pages={124093--124113},
  year={2024}
}

@article{miao2024inform,
  title={Inform: Mitigating reward hacking in rlhf via information-theoretic reward modeling},
  author={Miao, Yuchun and Zhang, Sen and Ding, Liang and Bao, Rong and Zhang, Lefei and Tao, Dacheng},
  journal={Advances in Neural Information Processing Systems},
  volume={37},
  pages={134387--134429},
  year={2024}
}

@inproceedings{zhang2024forget,
  title={Forget-me-not: Learning to forget in text-to-image diffusion models},
  author={Zhang, Gong and Wang, Kai and Xu, Xingqian and Wang, Zhangyang and Shi, Humphrey},
  booktitle={Proceedings of the IEEE/CVF conference on computer vision and pattern recognition},
  pages={1755--1764},
  year={2024}
}

@article{wu2024erasediff,
  title={Erasediff: Erasing data influence in diffusion models},
  author={Wu, Jing and Le, Trung and Hayat, Munawar and Harandi, Mehrtash},
  journal={arXiv preprint arXiv:2401.05779},
  year={2024}
}

@article{schuhmann2022laion,
  title={Laion-5b: An open large-scale dataset for training next generation image-text models},
  author={Schuhmann, Christoph and Beaumont, Romain and Vencu, Richard and Gordon, Cade and Wightman, Ross and Cherti, Mehdi and Coombes, Theo and Katta, Aarush and Mullis, Clayton and Wortsman, Mitchell and others},
  journal={Advances in neural information processing systems},
  volume={35},
  pages={25278--25294},
  year={2022}
}

@article{ouyang2022training,
  title={Training language models to follow instructions with human feedback},
  author={Ouyang, Long and Wu, Jeffrey and Jiang, Xu and Almeida, Diogo and Wainwright, Carroll and Mishkin, Pamela and Zhang, Chong and Agarwal, Sandhini and Slama, Katarina and Ray, Alex and others},
  journal={Advances in neural information processing systems},
  volume={35},
  pages={27730--27744},
  year={2022}
}

@article{shao2024deepseekmath,
  title={Deepseekmath: Pushing the limits of mathematical reasoning in open language models},
  author={Shao, Zhihong and Wang, Peiyi and Zhu, Qihao and Xu, Runxin and Song, Junxiao and Bi, Xiao and Zhang, Haowei and Zhang, Mingchuan and Li, YK and Wu, Yang and others},
  journal={arXiv preprint arXiv:2402.03300},
  year={2024}
}

@article{rafailov2023direct,
  title={Direct preference optimization: Your language model is secretly a reward model},
  author={Rafailov, Rafael and Sharma, Archit and Mitchell, Eric and Manning, Christopher D and Ermon, Stefano and Finn, Chelsea},
  journal={Advances in neural information processing systems},
  volume={36},
  pages={53728--53741},
  year={2023}
}

@misc{rapidata_flux2pro_t2i_human_preference_2025,
  author    = {Rapidata},
  title     = {Flux-2-pro\_t2i\_human\_preference},
  year      = {2025},
  publisher = {Hugging Face},
  url       = {https://huggingface.co/datasets/Rapidata/Flux-2-pro_t2i_human_preference},
  version   = {ef2e83f},
  note      = {Accessed: 2025-12-23}
}

@article{lee2023aligning,
  title={Aligning text-to-image models using human feedback},
  author={Lee, Kimin and Liu, Hao and Ryu, Moonkyung and Watkins, Olivia and Du, Yuqing and Boutilier, Craig and Abbeel, Pieter and Ghavamzadeh, Mohammad and Gu, Shixiang Shane},
  journal={arXiv preprint arXiv:2302.12192},
  year={2023}
}

@inproceedings{xu2024instructions,
  title={Instructions as backdoors: Backdoor vulnerabilities of instruction tuning for large language models},
  author={Xu, Jiashu and Ma, Mingyu and Wang, Fei and Xiao, Chaowei and Chen, Muhao},
  booktitle={Proceedings of the 2024 Conference of the North American Chapter of the Association for Computational Linguistics: Human Language Technologies (Volume 1: Long Papers)},
  pages={3111--3126},
  year={2024}
}

@inproceedings{hessel2021clipscore,
  title={Clipscore: A reference-free evaluation metric for image captioning},
  author={Hessel, Jack and Holtzman, Ari and Forbes, Maxwell and Le Bras, Ronan and Choi, Yejin},
  booktitle={Proceedings of the 2021 conference on empirical methods in natural language processing},
  pages={7514--7528},
  year={2021}
}

@article{heusel2017gans,
  title={Gans trained by a two time-scale update rule converge to a local nash equilibrium},
  author={Heusel, Martin and Ramsauer, Hubert and Unterthiner, Thomas and Nessler, Bernhard and Hochreiter, Sepp},
  journal={Advances in neural information processing systems},
  volume={30},
  year={2017}
}

@inproceedings{zhang2018unreasonable,
  title={The unreasonable effectiveness of deep features as a perceptual metric},
  author={Zhang, Richard and Isola, Phillip and Efros, Alexei A and Shechtman, Eli and Wang, Oliver},
  booktitle={Proceedings of the IEEE conference on computer vision and pattern recognition},
  pages={586--595},
  year={2018}
}

@article{nie2025v,
  title={V-Attack: Targeting Disentangled Value Features for Controllable Adversarial Attacks on LVLMs},
  author={Nie, Sen and Zhang, Jie and Yan, Jianxin and Shan, Shiguang and Chen, Xilin},
  journal={arXiv preprint arXiv:2511.20223},
  year={2025}
}

@article{kim2026improving,
  title={Improving Text-to-Image Generation with Intrinsic Self-Confidence Rewards},
  author={Kim, Seungwook and Cho, Minsu},
  journal={arXiv preprint arXiv:2603.00918},
  year={2026}
}

@misc{verlomni_github,
  title  = {{VeRL-Omni: Easy, Fast, and Stable RL Training for Diffusion and Omni-Modality Models}},
  author = {Yongxiang Huang and Cheung Kawai and Jingan Zhou and Yingshu Chen and {openYuanrong Team} and Xibin Wu},
  year={2026},
  howpublished={\url{https://github.com/verl-project/verl-omni}},
  note={Accessed: 2026-09-02}
}

@misc{compvis_safety_checker,
  author       = {{CompVis}},
  title        = {{Stable Diffusion Safety Checker}},
  url          = {https://huggingface.co/CompVis/stable-diffusion-safety-checker},
  note         = {Accessed: 2026-09-07}
}

@inproceedings{schramowski2022can,
  author       = {Schramowski, Patrick and Tauchmann, Christopher and Kersting, Kristian},
  title        = {Can Machines Help Us Answering Question 16 in Datasheets, and In Turn Reflecting on Inappropriate Content?},
  booktitle    = {Proceedings of the ACM Conference on Fairness, Accountability, and Transparency (FAccT)},
  year         = {2022},
  doi          = {10.1145/3531146.3533192}
}

@inproceedings{schramowski2023safe,
  author       = {Schramowski, Patrick and Brack, Manuel and Deiseroth, Bj{\"o}rn and Kersting, Kristian},
  title        = {Safe Latent Diffusion: Mitigating Inappropriate Degeneration in Diffusion Models},
  booktitle    = {Proceedings of the IEEE/CVF Conference on Computer Vision and Pattern Recognition (CVPR)},
  pages        = {22522--22531},
  year         = {2023},
  doi          = {10.1109/CVPR52729.2023.02157}
}

@misc{alibaba_qwen37plus_2026,
  author       = {{Alibaba Cloud}},
  title        = {{Qwen3.7-Plus}},
  year         = {2026},
  url          = {https://www.alibabacloud.com/help/en/model-studio/qwen3-7-plus},
  note         = {Accessed: 2026-09-07}
}

@misc{openai_gpt56terra_2026,
  author       = {{OpenAI}},
  title        = {{GPT-5.6 Terra Model}},
  year         = {2026},
  url          = {https://developers.openai.com/api/docs/models/gpt-5.6-terra},
  note         = {Accessed: 2026-09-07}
}

@misc{deepseek_v41flash_2026,
  author       = {{DeepSeek}},
  title        = {{DeepSeek-V4.1-Flash Release: Stronger, Faster, and More Accessible}},
  year         = {2026},
  url          = {https://api-docs.deepseek.com/zh-cn/news/news260910/},
  note         = {Accessed: 2026-09-10}
}

@misc{rapidata_docs_2026,
  author       = {Rapidata},
  title        = {Job Definition Parameter Reference: Quality Control},
  year         = {2026},
  url          = {https://docs.rapidata.ai/latest/job_definition_parameters/},
  note         = {Accessed: 2026-09-19}
}

@inproceedings{yang2025fifa,
  author       = {Yang, Yongjin and Kim, Sihyeon and Jung, Hojung and Bae, Sangmin and Kim, SangMook and Yun, Se-Young and Lee, Kimin},
  title        = {{Automated Filtering of Human Feedback Data for Aligning Text-to-Image Diffusion Models}},
  booktitle    = {International Conference on Learning Representations (ICLR)},
  year         = {2025}
}

@article{xu2026qwenimage2rl,
  author       = {Xu, Yixian and Gao, Kaiyuan and Chen, Yuxiang and Chen, Yilei and Tang, Zecheng and Liu, Zihao and Zhou, Zikai and Li, Deqing and others},
  title        = {{Qwen-Image-2.0-RL Technical Report}},
  journal      = {arXiv preprint arXiv:2606.27608},
  year         = {2026}
}

@inproceedings{liu2008isolation,
  author       = {Liu, Fei Tony and Ting, Kai Ming and Zhou, Zhi-Hua},
  title        = {{Isolation Forest}},
  booktitle    = {IEEE International Conference on Data Mining (ICDM)},
  pages        = {413--422},
  year         = {2008}
}

@article{tran2018spectral,
  title={Spectral signatures in backdoor attacks},
  author={Tran, Brandon and Li, Jerry and Madry, Aleksander},
  journal={Advances in neural information processing systems},
  volume={31},
  year={2018}
}

@inproceedings{hayase2021spectre,
  author       = {Hayase, Jonathan and Kong, Weihao and Somani, Raghav and Oh, Sewoong},
  title        = {{SPECTRE}: Defending Against Backdoor Attacks Using Robust Statistics},
  booktitle    = {International Conference on Machine Learning (ICML)},
  pages        = {4129--4139},
  year         = {2021}
}


\begin{IEEEbiography}[{\includegraphics[width=1in,height=1.25in,clip,keepaspectratio]{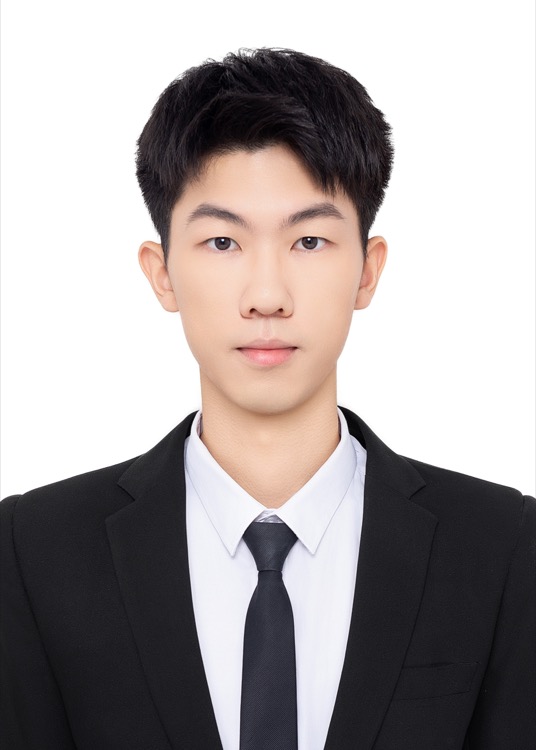}}]{Kaiwen Duan}
received the B.S. degree in information security from Wuhan University in 2025. He is currently pursuing the M.S. degree with Zhejiang University, Hangzhou, China. His research interests include the security of text-to-image generative models and large language models.
\end{IEEEbiography}

\begin{IEEEbiography}[{\includegraphics[width=1in,height=1.25in,clip,keepaspectratio]{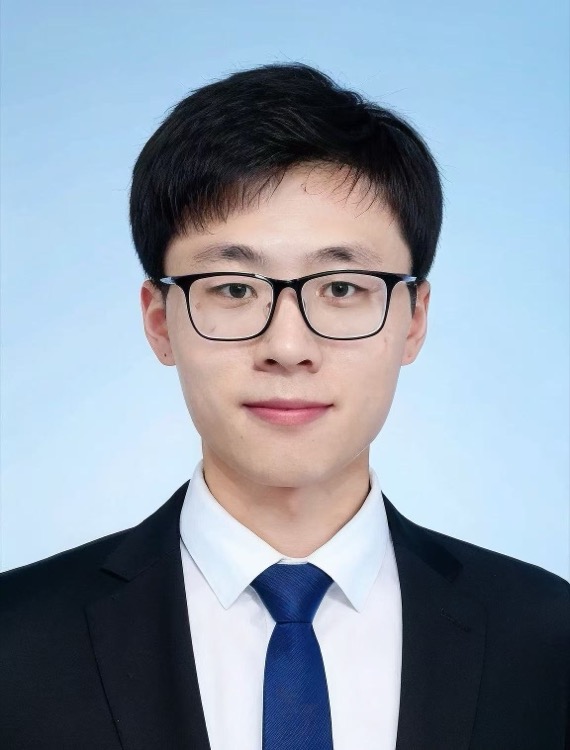}}]{Hongwei Yao}
is currently a Postdoctoral Fellow with the City University of Hong Kong, working with Prof. Cong Wang. He received the Ph.D. degree from the College of Computer Science and Technology, Zhejiang University, advised by Dr. Zhan Qin. His research focuses on trustworthy AI and the security and privacy of large language models.
\end{IEEEbiography}

\begin{IEEEbiography}[{\includegraphics[width=1in,height=1.25in,clip,keepaspectratio]{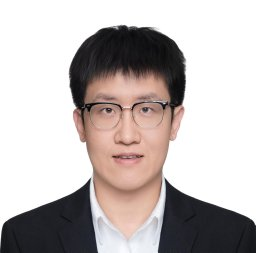}}]{Yufei Chen}
received the dual Ph.D. degree from Xi'an Jiaotong University and the City University of Hong Kong in 2024. He is currently a Research Assistant Professor with the Department of Computer Science, City University of Hong Kong. His research interests lie at the intersection of artificial intelligence and cybersecurity, with a focus on adversarial machine learning, security and privacy of AI systems, and privacy-preserving machine learning.
\end{IEEEbiography}

\begin{IEEEbiography}[{\includegraphics[width=1in,height=1.25in,clip,keepaspectratio]{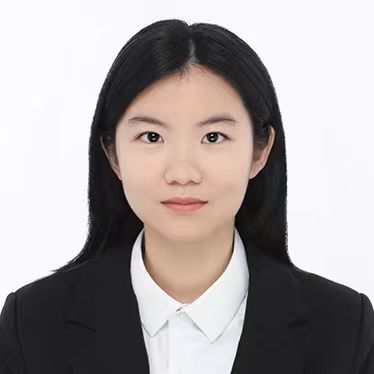}}]{Ziyun Li}
is a Postdoctoral Researcher with the KTH Royal Institute of Technology, Stockholm, Sweden, funded by the Wallenberg AI, Autonomous Systems and Software Program (WASP). She received the Ph.D. degree from the Hasso Plattner Institute, University of Potsdam, Germany, in 2024. Her research interests include generative models, uncertainty quantification, and AI security.
\end{IEEEbiography}

\begin{IEEEbiography}[{\includegraphics[width=1in,height=1.25in,clip,keepaspectratio]{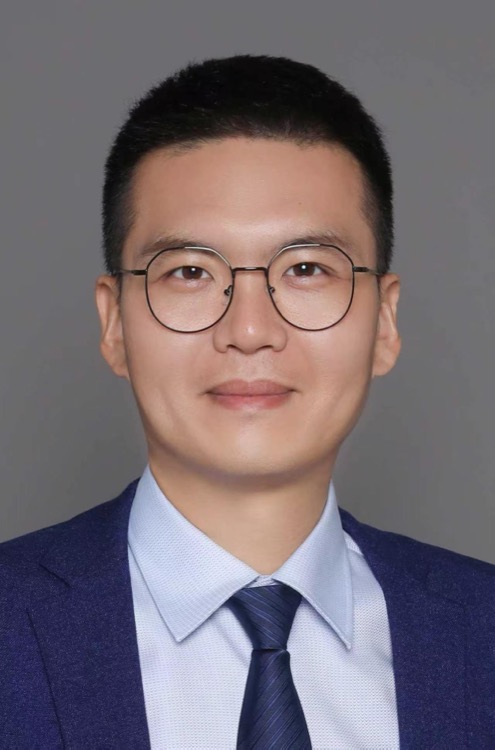}}]{Tong Qiao}
received the B.S. degree in electronic and information engineering from Information Engineering University, Zhengzhou, China, in 2009, the M.S. degree in communication and information system from Shanghai University, Shanghai, China, in 2012, and the Ph.D. degree from the Laboratory of Systems Modelling and Dependability, University of Technology of Troyes, Troyes, France, in 2016. He is currently a Full Professor with the School of Cyberspace, Hangzhou Dianzi University. His current research interests include media forensics, AI security, and data hiding. He has published over 80 peer-reviewed papers in journals and conferences.
\end{IEEEbiography}

\begin{IEEEbiography}[{\includegraphics[width=1in,height=1.25in,clip,keepaspectratio]{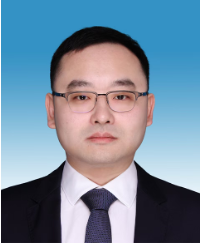}}]{Zhan Qin}
received the Ph.D. degree from the Department of Computer Science and Engineering, State University of New York at Buffalo, in 2017. He is currently an Associate Professor with both the College of Computer Science and Technology and the Institute of Cyberspace Research (ICSR), Zhejiang University, China. He was an Assistant Professor with the Department of Electrical and Computer Engineering, University of Texas at San Antonio. His current research interests include data security and privacy, secure computation outsourcing, artificial intelligence security, and cyber-physical security in the context of the Internet of Things. His works explore and develop novel security-sensitive algorithms and protocols for computation and communication in the general context of cloud and Internet devices.
\end{IEEEbiography}

\begin{IEEEbiography}[{\includegraphics[width=1in,height=1.25in,clip,keepaspectratio]{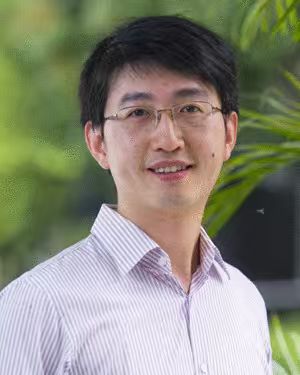}}]{Cong Wang}
(Fellow, IEEE) is currently a Chair Professor and the Head of the Department of Computer Science, College of Computing, City University of Hong Kong. His research encompasses data security and privacy, AI systems and security, and blockchain with decentralized applications. He has made prolific contributions to these fields, witnessed by more than 30,000 citations on Google Scholar. He is an HK RGC Research Fellow and a Founding Member of the Young Academy of Sciences of Hong Kong. He received multiple best paper awards, including the 2020 IEEE INFOCOM Test of Time Paper Award. He served as the Editor-in-Chief of the \textit{IEEE Transactions on Dependable and Secure Computing} from 2022 to 2025, a premier security journal under the IEEE Computer Society.
\end{IEEEbiography}

\end{document}